\pdfoutput=1
\documentclass[11pt]{article}

\usepackage[preprint]{acl}
\usepackage{times}
\usepackage{latexsym}
\usepackage[T1]{fontenc}
\usepackage[utf8]{inputenc}
\usepackage{microtype}
\usepackage{inconsolata}
\usepackage{graphicx}
\usepackage{booktabs}
\usepackage{makecell}
\usepackage{amsmath,amssymb}
\usepackage{xcolor}
\usepackage[inline]{enumitem}
\usepackage{adjustbox}
\usepackage{tikz}
\usetikzlibrary{arrows.meta}
\definecolor{gpblue}{HTML}{2A78D6}
\definecolor{gporange}{HTML}{EB6834}

\newcommand{\ex}[1]{\textit{#1}}
\newcommand{\cond}[1]{\textsc{#1}}
\newcommand{\ci}[2]{{\scriptsize\,(#1, #2)}}


\title{Language Models Keep Abandoned Analyses:\\ Residue and Adaptation after Garden-Path Sentences}

\author{Xiang Ding \\ Nagoya University \\ \texttt{xiang.ding.i8@s.mail.nagoya-u.ac.jp}}

\begin{document}
\maketitle

\begin{abstract}
Reading \ex{While the man hunted the deer ran into the woods} requires giving up the analysis in which \ex{the deer} is the object of \ex{hunted}.
People who have just read such a sentence still produce more transitive structures, so the abandoned analysis primes later language.
We ask whether language models keep using an analysis they have abandoned, and what in the model carries it.
In 24 open models from 124M to 70B parameters, a garden-path sentence leaves two traces in the context.
A residue of the abandoned analysis raises later object attachment, by about three quarters of the effect of a sentence in which the noun phrase really is an object, and it appears in sampled text, on held-out and human-normed items and for a second type of garden path.
An adaptation toward the resolved structure scales with the sentence's disambiguation error and acts mainly on comprehension-like measures.
Activation patching separates the two.
Later text reads the residue from the role information in the noun phrase's states, which were computed under the object analysis before the disambiguation, and replacing these states removes almost all of the residue in the 32B and 70B models.
The adaptation is read from the states that processed the disambiguation.
Overridden statements show the same persistence (a corrected value keeps pulling later predictions in 8 of 9 models), and in most models the pull is carried by the positions after the statement.
Finally, the human paradigm transfers only with controls for punctuation, because the models also write garden-path sentences themselves, which reverses the paradigm's raw statistic.
\end{abstract}

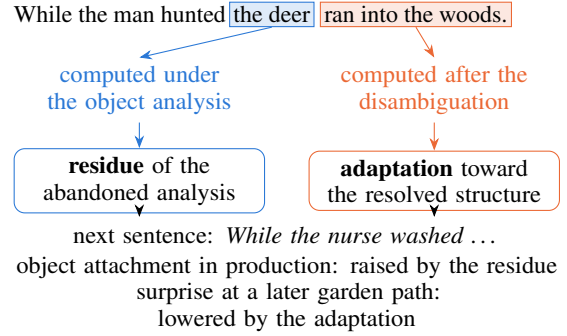
\begin{figure}[t]
\centering
\begin{tikzpicture}[font=\footnotesize, >=Stealth, box/.style={text width=3.1cm, align=center, rounded corners, inner sep=3pt}]
\node[anchor=west, inner sep=1pt] (pre) at (0,0) {While the man hunted};
\node[anchor=west, inner sep=1.5pt, fill=gpblue!15, draw=gpblue] (np) at (pre.east) {the deer};
\node[anchor=west, inner sep=1.5pt, fill=gporange!15, draw=gporange] (mc) at ([xshift=2pt]np.east) {ran into the woods.};
\node[box, text=gpblue] (npl) at (1.7,-1.0) {computed under the object analysis};
\node[box, text=gporange] (mcl) at (5.6,-1.0) {computed after the disambiguation};
\draw[->, gpblue] (np.south) -- (npl.north);
\draw[->, gporange] (mc.south) -- (mcl.north);
\node[box, draw=gpblue] (r) at (1.7,-2.2) {\textbf{residue} of the abandoned analysis};
\node[box, draw=gporange] (a) at (5.6,-2.2) {\textbf{adaptation} toward the resolved structure};
\draw[->, gpblue] (npl) -- (r); \draw[->, gporange] (mcl) -- (a);
\node[text width=7.4cm, align=center] (t) at (3.65,-3.5) {next sentence: \emph{While the nurse washed \ldots}\\[1pt]
object attachment in production: raised by the residue\\ surprise at a later garden path: lowered by the adaptation};
\draw[->] (r.south) -- (r.south |- t.north); \draw[->] (a.south) -- (a.south |- t.north);
\end{tikzpicture}
\caption{Two traces of a garden-path sentence in a causal language model.
Later text reads a residue of the abandoned object analysis from the states of \ex{the deer}, which were computed before the disambiguation (\S\ref{sec:mechanism}).
The main-clause states carry an adaptation toward the resolved structure.
The two traces push later processing in opposite directions, and their balance differs between production-like and comprehension-like measures (\S\ref{sec:twoforces}).}
\label{fig:schema}
\end{figure}

\section{Introduction}

Garden-path sentences make a reader build an analysis and then give it up.
In \ex{While the man hunted the deer ran into the woods}, the noun phrase \ex{the deer} is first attached as the object of \ex{hunted}, and the verb \ex{ran} reveals that it is the subject of the main clause \citep{frazier1982}.
In language models (LMs), which have become a standard test bed for such sentences, surprisal at the disambiguating word predicts that garden-path effects occur but underestimates their size \citep{futrell2019,vanschijndel2021,arehalli2022,huang2024}.
Asked \ex{Did the man hunt the deer?}, they often answer yes, like human readers who keep part of the initial interpretation \citep{christianson2001,li2024incremental,amouyal2025}.
These studies examine the garden-path sentence itself, either the difficulty of its disambiguation or the accuracy of its final interpretation.

The abandoned analysis can also shape what comes after the sentence.
\citet{vangompel2006} had participants read a garden-path sentence and then complete a fragment such as \ex{While the nurse washed}.
Completions were more often transitive after the garden path than after a version disambiguated by a comma (33.6\% against 29.1\% in their first experiment), so the object analysis that had been ruled out still primed later production.
For LMs this question has a specific architectural form.
A causal transformer computes the representation of each token when the token is read, and later tokens retrieve these representations through attention.
When \ex{ran} arrives, the states of \ex{the deer} have already been computed under the object analysis, and they stay in the context unchanged.
The states computed after the disambiguation, in contrast, can encode the corrected analysis.
Whether later text is shaped by the abandoned analysis therefore depends on which of these states it reads, and this cannot be decided from the architecture alone.

Studies of structural priming in LMs have so far not been able to address this question.
LMs reuse the structure of a preceding sentence \citep{sinclair2022,prasad2019,michaelov2023,pucci2026}, and they do so more after less expected primes, an inverse-frequency effect that \citet{zhou2025icl} take as evidence that in-context learning is error-driven \citep[see also][]{jumelet2024}.
In human research the same effect separates two accounts of priming.
Residual activation holds that a processed structure stays active and is reused because it is active \citep{pickering1998}.
Error-driven learning holds that a prime adjusts expectations in proportion to its prediction error \citep{chang2006,jaeger2013}.
With unambiguous primes, both accounts predict priming of the prime's structure, and only the scaling with prediction error tells them apart.

Garden-path primes add a second contrast.
For the abandoned object analysis, residual activation predicts reuse that grows with how strongly the analysis was built.
Error-driven learning predicts the opposite tendency, because the error occurs when the object analysis fails, and learning from it should favour the resolved structure in which \ex{the deer} starts a new clause.
Human data show only the net result.
LMs let us measure the effect of a garden-path prime on later text, relate it to the strength of the abandoned analysis and to the disambiguation error, and locate in the context what carries each part.

We pair garden-path primes from five published item sets with target fragments and measure how a prime changes a production-like quantity, the preference for attaching the target's noun phrase as an object, and a comprehension-like quantity, the surprisal of the disambiguating word in a garden-path target.
Across 24 models we compare primes matched for punctuation and verb, fix the predictions of each confirmatory experiment before collecting its data, and use activation patching to locate the effects inside the prime.

A garden-path prime raises later object attachment by about three quarters of the effect of a prime whose noun phrase really is an object, and most of this residue requires a noun phrase that could have been an object.
The same prime produces an adaptation toward the resolved structure that grows with its disambiguation error and acts mainly on the comprehension-like measure, and patching shows that later text reads the two from different positions of the prime.
To see how far the persistence extends, we also test statements that a later sentence corrects.
The overridden value keeps influencing later predictions, but through a different route than the garden path.

Our contributions are as follows.
(1) We show that LMs carry an abandoned syntactic analysis into later production, across 24 models, three item sets and two types of garden path, with controls that rule out punctuation, verb bias and local string overlap as its source.
(2) We show that the same prime also produces an adaptation toward the resolved analysis, that the two effects scale with different properties of the prime (the strength of the abandoned analysis and the disambiguation error), and that their balance differs between production-like and comprehension-like measures.
(3) We show by activation patching, probing and a role-direction intervention that later text reads the abandoned analysis from the role information in noun-phrase states computed before the disambiguation, even though the states after it encode the reanalysis.
(4) We show that overridden statements also persist, in most models through later positions, and that the human paradigm's statistic reverses in LMs unless punctuation is controlled.

\section{Related Work}

\paragraph{Garden paths in language models.}
Garden-path sentences were among the first constructions used to ask whether neural LMs track syntactic state \citep{futrell2019}.
Surprisal from such models predicts that the effects exist but underestimates their size and misorders constructions \citep{vanschijndel2021,arehalli2022,huang2024}.
Recent work relates this gap to the number of parses a model keeps alive \citep{timkey2026} or closes it with suitably trained models \citep{yoshida2026}.
Comprehension studies ask whether the final interpretation is correct, and find that LMs often accept the initial misreading, with plausibility and verb type modulating this as in humans \citep{li2024incremental,amouyal2025}, and difficulty with such structures persists across model scale \citep{amouyal2026}.
\citet{hanna2025} found features for both readings of a garden-path prefix in small models and showed that the circuit that answers comprehension questions hardly reuses them.
None of this work asks what the abandoned analysis does to later text.

\paragraph{Structural priming and adaptation.}
Speakers reuse structures they have recently processed \citep{bock1986,pickering1998}.
Residual activation and error-driven learning \citep{chang2006} are the two main accounts, and computational models combine short-lived activation with longer-lived learning \citep{reitter2011}, in line with a lexical boost that decays while abstract priming persists \citep{hartsuiker2008}.
Less expected primes prime more \citep{jaeger2013}, and readers adapt to the main-verb/reduced-relative garden path with exposure \citep{fine2013}, although such rapid adaptation is small and hard to replicate \citep{harringtonstack2018,prasad2021}.
\citet{vangompel2006} found priming of the abandoned analysis of NP/Z garden paths, and \citet{fujita2020} found that unambiguous primes ease the reading of later garden paths without removing lingering misinterpretation.
In LMs, conditioning on a prime raises the probability of its structure in a target \citep{prasad2019,sinclair2022}, across languages \citep{michaelov2023} and in generation \citep{pucci2026}.
The effect grows for less expected primes and with lexical overlap \citep{jumelet2024,zhou2025icl}, and gradient updates on garden-path sentences improve the fit of RNN surprisal to human adaptation \citep{vanschijndel2018}.
Apart from gradient-based adaptation to garden paths \citep{vanschijndel2018}, these studies use unambiguous primes.

\paragraph{Overridden information in context.}
Information that later context overrides can continue to affect LMs.
Transformer models sometimes refer to discourse entities that a sentence did not introduce \citep{schuster2022}, and models rarely retract their own wrong answers \citep{yang2025retraction} or correct themselves without external feedback \citep{kamoi2024}.
Later tokens retrieve earlier bindings by attending back to the positions where they were encoded \citep{prakash2026}, and activation patching, which replaces internal states with states from a counterfactual input, is the standard tool for locating such information \citep{vig2020causal,meng2022locating,zhang2024patching}.
In human sentence processing, lingering misinterpretations have been attributed to competing representations of the two analyses \citep{slattery2013} and to shallow, good-enough interpretation \citep{ferreira2003}.
Garden paths offer a case of overriding in which the overridden analysis, its strength and the point at which it is overridden are all known.

\section{Experimental Setup}
\label{sec:setup}

\paragraph{Materials.}
An item family is one garden-path sentence with its controls.
The discovery set contains 93 NP/Z families, 69 from \citet{amouyal2025} and 24 from the SAP benchmark \citep{huang2024}.
A held-out set of 48 families combines 24 items from \citet{cehakova2025} with 24 from SyntaxGym \citep{gauthier2020}, and was used only after the analyses had been fixed on the discovery set.
A third set contains the 42 optionally transitive items of \citet{christianson2001}, each with a short and a long ambiguous region and a plausible and an implausible object.
Each family provides the primes in Table~\ref{tab:conds}.
\cond{gp} is the garden-path sentence and \cond{cue} its version with a comma after the verb.
\cond{tb} replaces the main-clause subject with a pronoun, so that the noun phrase remains the object.
\cond{nb} and \cond{nc} replace the ambiguous verb with a strictly intransitive one, without and with the comma.

\begin{table}[t]
\centering\small
\setlength{\tabcolsep}{3pt}
\begin{tabular}{lp{5.7cm}}
\toprule
Prime & Example\\
\midrule
\cond{gp} & While the man hunted the deer ran into the woods.\\
\cond{cue} & While the man hunted, the deer ran into the woods.\\
\cond{tb} & While the man hunted the deer it ran into the woods.\\
\cond{nb} / \cond{nc} & While the man slept(,) the deer ran into the woods.\\
\cond{nom} & While the man hunted they ran into the woods.\\
\cond{pass} & While the man was hunted the deer ran into the woods.\\
\midrule
Target & Quantity\\
\midrule
$L$ & $\log P(\text{\ex{\ldots washed the baby,}}) - \log P(\text{\ex{\ldots washed, the baby}})$\\
$S$ & $-\log P(\text{\ex{cried}} \mid \text{\ex{While the nurse washed the baby}})$\\
$G$ & sampled continuation of \ex{While the nurse washed}\\
\bottomrule
\end{tabular}
\caption{Prime conditions and target quantities, illustrated with one prime and one target family. \cond{nom} and \cond{pass} test punctuation and local coherence (\S\ref{sec:residue}).}
\label{tab:conds}
\end{table}

\paragraph{Targets and measures.}
Each prime is followed by a target from every other family that shares no content word with it, which removes lexical repetition and gives each discovery prime 78 to 92 targets.
For a target family with subject $s$, verb $v$ and noun phrase $n$, the production-like measure is
\begin{equation}
L = \log P(\text{\ex{While}}\ s\ v\ n\text{,}) - \log P(\text{\ex{While}}\ s\ v\text{,}\ n),
\end{equation}
where both probabilities are conditioned on the prime and the two strings contain the same words.
$L$ is the log-odds of attaching $n$ as the object of $v$ rather than starting a new clause with it.
The comprehension-like measure $S$ is the surprisal of the first word of the target's main clause after the comma-less fragment \ex{While}~$s$~$v$~$n$, a quantity that contains no punctuation choice.
A high $S$ means that the model expected $n$ to be an object.
For sampled production, the model continues \ex{While}~$s$~$v$, and a rule system over part-of-speech tags codes whether the continuation attaches a noun phrase as an object (Appendix~\ref{app:coding}).

\paragraph{The residue contrast.}
Comparing \cond{gp} with \cond{cue} confounds the abandoned analysis with the comma, and comparing \cond{gp} with \cond{nb} confounds it with the verb.
For any measure $X$ we use
\begin{equation}
A_X = \big(X_{\cond{gp}} - X_{\cond{nb}}\big) - \big(X_{\cond{cue}} - X_{\cond{nc}}\big),
\label{eq:A}
\end{equation}
where $X_c$ is the mean of $X$ over prime families and targets after primes of condition $c$.
Within each bracket the two primes have the same punctuation, and across brackets the same verbs.
$A_X$ is therefore the part of the prime's effect that requires both a missing comma and a verb that can take an object, the condition under which the object analysis is built and then abandoned.
The same contrast with \cond{tb} in place of \cond{gp}, $A_X^{\cond{tb}}$, measures the effect of a noun phrase that really is an object, and we report $r=A_X/A_X^{\cond{tb}}$ as the size of the residue relative to it.

\paragraph{Models and inference.}
We use 24 open-weight causal LMs from the GPT-2, Pythia, Llama, Mistral, Gemma and Qwen families, base and instruction-tuned, between 124M and 70B parameters and including one mixture-of-experts model (Appendix~\ref{app:models}).
22 of them enter every priming experiment, and two more enter the \citet{christianson2001} set.
Scoring uses exact log-probabilities without chat templates, and sampling uses temperature 1 without truncation.
Intervals come from a two-way bootstrap that resamples prime families and target families independently.
For averages across models, each bootstrap draw applies the same resampled families to every model, which keeps the dependence that arises because all models see the same items.
For every confirmatory experiment we fixed predictions and analysis in a dated record before collecting its data, and Appendix~\ref{app:prereg} lists every outcome.

\begin{figure*}[t]
\centering
\includegraphics[width=\textwidth]{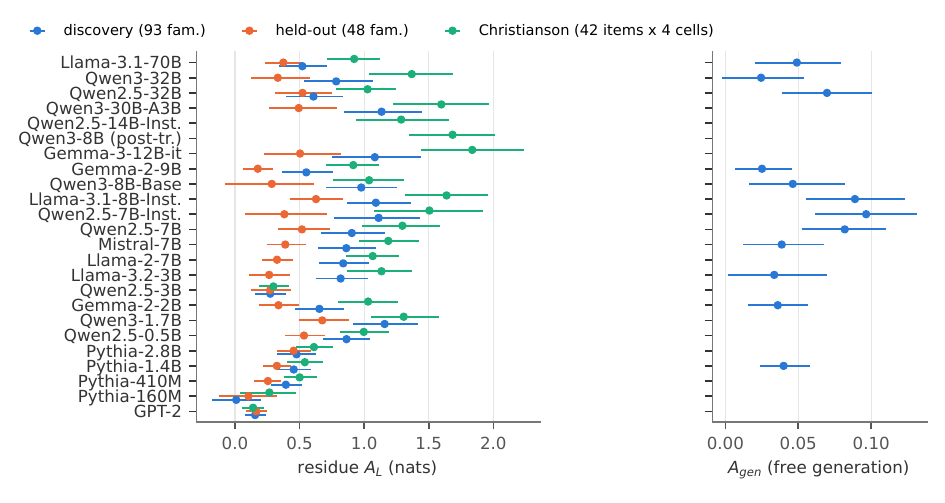}
\caption{The residue of the abandoned object analysis. Left: $A_L$ (Eq.~\ref{eq:A}) per model on the discovery, held-out and \citet{christianson2001} item sets. Right: the same contrast on sampled completions, as a difference in the proportion of transitive continuations. Error bars are 95\% intervals from a two-way bootstrap over prime and target families. Models are ordered by size.}
\label{fig:residue}
\end{figure*}

\section{The Abandoned Analysis Persists}
\label{sec:residue}

\paragraph{The residue in production.}
After a garden-path prime, models prefer object attachment in the target more than the controls predict (Figure~\ref{fig:residue}).
On the discovery set $A_L$ is positive in all 22 models and excludes zero in 21, with Pythia-160M as the exception.
Averaged over models it is 0.72 nats, which is 76\% of $A_L^{\cond{tb}}$, the effect of a prime whose noun phrase really is the object.
The residue replicates on the held-out families, where it is positive in all 22 models (20 with intervals above zero) with a median of 0.36 nats, and on the items of \citet{christianson2001}, where it is positive in all 24 models with a median of 1.05 nats.

\paragraph{Punctuation habits.}
A comma-less prime is unusual in written text, and a model could learn from it that this writer omits commas after verbs that can take objects.
Such a habit would raise $L$ without any object analysis.
The \cond{nom} prime keeps the verb and the missing comma of \cond{gp} but replaces the noun phrase with the nominative pronoun \ex{they}, which cannot be an object.
If the residue were a punctuation habit, \cond{gp} and \cond{nom} would prime alike.
Instead, the contrast of Equation~\ref{eq:A} with \cond{nom} and its intransitive counterpart in place of \cond{cue} and \cond{nc} is positive in all 22 models (21 significant, 0.47 nats on average), and across families it grows with the model's expectation of an object after the ambiguous verb in all 22 models.
The case contrast is two thirds of the comma contrast (0.47 against 0.72 nats), so most of the residue requires a noun phrase that could be an object, and the rest follows from the object expected after the verb.

\paragraph{Local string overlap.}
The string \ex{hunted the deer} might prime \ex{washed the baby} by local coherence, as it disrupts human reading even where the grammar excludes an object reading \citep{tabor2004}.
The \cond{pass} prime contains the same string inside a passive clause, where \ex{the deer} cannot be an object.
Its residue has a median of 12\% of the \cond{gp} residue across 22 models, reliably negative in seven, mostly small, models and reliably positive in seven larger ones.
The ratio is higher in the strongest models, 0.45 in Llama-3.1-70B and above 0.5 in the two Qwen2.5-7B models.
In these models a locally computed object reading of the string also leaves a trace.

\paragraph{Sampled production.}
We sampled two continuations of every target fragment after every prime in 12 models, about 57,000 prompts per model.
The residue appears in the proportion of transitive continuations in 11 of 12 models with the pre-specified coding and in all 12 with a refined coding, by 5.0 percentage points on average (Figure~\ref{fig:residue}, right).
Its size across models followed $A_L$ in the eight models available when we specified this prediction (Spearman $\rho=0.81$), and the relation is weaker across all twelve ($\rho=0.48$).
The human raw contrast between garden-path and comma primes was 4.5 points \citep{vangompel2006}, a statistic that cannot be compared directly with LMs (\S\ref{sec:transfer}).

\paragraph{A second garden path.}
In an NP/S garden path (\ex{The student forgot the solution was in the back of the book}) the noun phrase is also first analysed as an object.
We built the same contrast for 64 NP/S families, with \ex{that} as the disambiguating cue and a verb that takes only a clausal complement (\ex{hoped}) as the control, and measured it on the NP/Z targets.
The residue is positive in 19 of 22 models (sign test $p<0.001$) with a pooled value of 0.03 nats.
It is smaller than for NP/Z primes because the prime and target structures differ, and even a real object in an NP/S prime moves the NP/Z targets only by 0.03 to 0.20 nats.

\section{Two Effects with Opposite Signs}
\label{sec:twoforces}

On the comprehension-like measure the residue is small on average (0.05 nats over 22 models) and varies in sign, significantly positive in five models and significantly negative in four.
A negative value means that after a garden-path prime the model expects an object less than the controls predict.
This is what an adaptation toward the resolved structure would produce, and it suggests that a second effect of the prime offsets the residue.

\paragraph{Commitment and error.}
The two accounts of priming tie the effect of a prime to different properties.
Residual activation depends on how strongly the abandoned analysis was built, which we measure as the model's expectation of an object after the ambiguous verb, $\log P(\text{\ex{the}}) - \log P(\text{\ex{,}})$.
Error-driven learning depends on the error at the disambiguation, which we measure as the garden-path strength of the prime, the surprisal of its disambiguating word without the comma minus that with the comma.
The two predictors are only weakly correlated (median $r=0.22$).
In a regression of the family-level residue on both (Figure~\ref{fig:slopes} in Appendix~\ref{app:stats}), object expectation raises the residue on both measures in all 22 models.
Garden-path strength lowers the residue on $S$ in 21 of 22 models, in all 22 when the family's adaptation to the missing comma enters as a covariate (19 significant), whereas on $L$ it raises the residue in most models.
The error effect is specific to garden-path resolutions: on the comma version of the same target, where the disambiguating word is expected, garden-path strength has little effect (median $t=-0.6$, against $-2.8$ on $S$), so the adaptation is not a general loss of confidence after a surprising prime.

\paragraph{Causal tests.}
Garden-path strength varies between families together with other properties.
To manipulate it within a family, we kept each prime up to the noun phrase fixed, which fixes the object analysis, and varied only its continuation, which changes the surprisal at the disambiguation.
More surprising disambiguations reduce the residue on $S$, with $t$ values between $-4.1$ and $-6.4$ in the three models whose between-family analysis had shown an error effect, and near zero in the two models in which it had not.
Surprisal summed over the whole continuation predicts little, so the relevant error is the one at the disambiguation point.
On the items of \citet{christianson2001}, where plausibility changes only the main-clause continuation, the same within-family relation is negative in all ten models that pass the comprehension controls of \S\ref{sec:transfer}.
The residue is tested causally by the patching experiments of \S\ref{sec:mechanism} (see Appendix~\ref{app:e014} for a behavioural manipulation).

\paragraph{Production and comprehension.}
The residue dominates the production-like measure, and the adaptation shows mainly in the comprehension-like one.
We tested this asymmetry prospectively on the items of \citet{christianson2001} with the 16 models available at that time.
The residue on $L$ was positive in all 16, and the error relation was more negative on $S$ than on $L$ in 14.
Human studies examine these two effects in separate paradigms \citep{vangompel2006,fine2013}, whereas in LMs they arise from the same prime.

\paragraph{Other garden paths.}
We tested the two item-level relations on $S$ for two further constructions with targets of the same construction.
For NP/S garden paths neither relation appeared.
For main-verb/reduced-relative garden paths the error relation appeared in six of nine models and the commitment relation in none.
NP/S reanalysis keeps the noun phrase inside the verb's complement, whereas NP/Z and main-verb/reduced-relative reanalysis remove a committed attachment \citep{sturt1996}, which may explain why only the latter produce an adaptation, while the NP/S residue of \S\ref{sec:residue} shows that the abandoned object analysis persists in both.

\section{Where the Residue Lives}
\label{sec:mechanism}

In the garden-path prime, the states of \ex{the deer} are identical to those in the transitive prime \cond{tb}, since the two share every preceding token.
Later tokens can read these states directly, or the main-clause states, which were computed after the disambiguation.
If the residue is read from the noun-phrase states, replacing them with states computed under the subject analysis should remove it, while replacing the main-clause states should not.

\paragraph{Patching design.}
At every layer, the residual-stream states of the noun-phrase tokens in \cond{gp} are replaced with the states of the same tokens in \cond{cue}, where the comma has already made the noun phrase a subject, and the rest of the prime is recomputed.
The target then attends to a mixture of the original and this counterfactual prime: counterfactual keys and values at the noun phrase (NP), at the main clause \ex{ran into the woods.} (MC), or at both, and original ones elsewhere.
The same procedure applied to \cond{nb} with \cond{nc} as donor estimates what the replacement does by importing information about the comma, and Equation~\ref{eq:A} subtracts it.
We report the retained residue $\rho = A_L^{\text{patched}}/A_L$ and the change $\Delta = A_S^{\text{patched}} - A_S$ for nine models between 1.4B and 70B parameters.

\begin{figure*}[t]
\centering
\includegraphics[width=\textwidth]{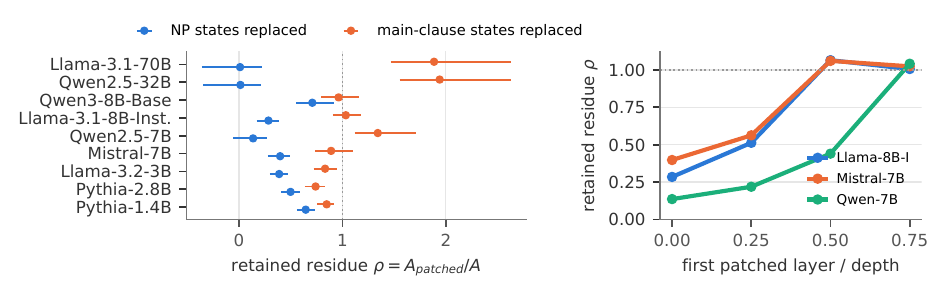}
\caption{Patching the garden-path prime. Left: residue on $L$ retained when the target reads counterfactual subject-analysis states at the noun phrase (blue) or at the main clause (orange). Right: noun-phrase states replaced only from a given fraction of the network's depth onward. Values for both measures, with intervals, are in Appendix~\ref{app:patching}.}
\label{fig:mech}
\end{figure*}

\paragraph{The residue is read from the noun phrase.}
Replacing the noun-phrase states leaves 0.01 to 0.71 of the residue (pooled 0.37), and nothing beyond the interval bounds of 0.21 and 0.22 in Qwen2.5-32B and Llama-3.1-70B (Figure~\ref{fig:mech}, left).
Replacing the main-clause states leaves it largely intact (pooled 1.01), and in every model the noun phrase carries more of the residue than the main clause.
With \cond{nb} as donor, where \ex{the deer} follows an intransitive verb without a comma, the retained fraction is still 0.04 to 0.47 in the seven models tested, so the effect is not imported information about the comma.
Replacing the noun-phrase states only in the upper half of the network has no effect in Llama-3.1-8B-Instruct and Mistral-7B, and only in the upper quarter none in Qwen2.5-7B (Figure~\ref{fig:mech}, right).
Later text thus reads the abandoned analysis from the noun phrase's keys and values in the lower and middle layers.

\paragraph{What the noun-phrase states carry.}
Replacing the noun-phrase states with states of the same tokens computed under the object analysis in another context (\ex{It was a quiet morning.} before the prime) leaves 70\% to 111\% of the residue in seven models up to 70B, so the effect of the replacement lies in the analysis it removes.
The states encode that analysis as a role.
A linear probe trained only on noun phrases without a comma, object (\cond{tb}) against subject (\cond{nb}), classifies object and subject noun phrases that follow a comma (\cond{tc}, \cond{cue}) with 96\% to 99\% accuracy in all seven models.
Moving the states along the probe direction toward the subject class removes more of the residue than random directions in three of seven models and less than replacing the whole states in six (Appendix~\ref{app:patching}), so the abandoned analysis is spread over more than one direction of the noun phrase's representation.

\section{Beyond Syntax: Overridden Statements}
\label{sec:facts}

Does content that a context explicitly overrides also keep influencing an LM?
Unlike conflicts between a context and a model's parametric knowledge \citep{longpre2021,xie2024conflict}, both the overridden and the correct value here come from the context.
We built 200 items from ten frames (for example \ex{The meeting is on Thursday. Correction: The meeting is on Monday.}) and measured, after the probe \ex{To confirm, the meeting is on}, how much more likely the overridden value is than the correct one.
A control replaces the overridden statement with the same statement about another entity (\ex{The dinner is on Thursday.}), which matches exposure to the overridden value.
Relative to this control, the overridden value pulls later predictions toward itself by 0.6 to 2.3 nats in eight of nine models up to 70B, in nine or ten of the ten frames each.
Qwen2.5-7B-Instruct is the exception and moves away from a corrected value.

Attention knockout shows where the answer finds the overridden value.
Blocking the answer's attention to the correcting sentence raises the pull in all nine models, so the knockout works as intended.
In six models, blocking the answer's attention to the overridden statement leaves the pull intact, and blocking the access of all later positions to it removes the pull: the value reaches the answer through the positions after the statement, including the correction itself.
In Qwen2.5-7B, Qwen2.5-32B and Llama-3.1-70B, blocking the answer's own access already halves or removes the pull, so the answer reads the overridden statement directly, as later text reads the abandoned noun phrase in a garden path (Appendix~\ref{app:facts}).
Overridden content thus persists in both settings, and whether it is read from where it was stated or relayed by later positions depends on the model.

\section{Transferring the Human Paradigm}
\label{sec:transfer}

\begin{figure}[t]
\centering
\includegraphics[width=\columnwidth]{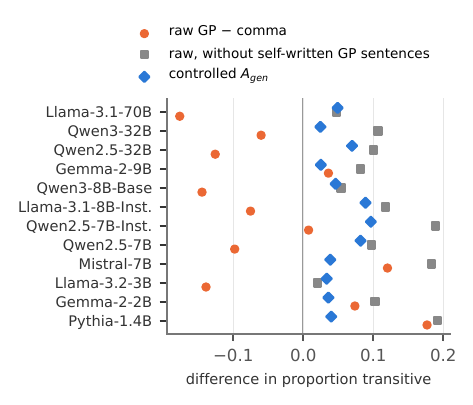}
\caption{Sampled production after garden-path primes. The raw contrast with comma primes used in the human study (orange) is negative in most Llama and Qwen models. Without the completions that are themselves comma-less garden-path sentences it is positive in every model (gray), and so is the controlled residue $A_{gen}$ (blue).}
\label{fig:generation}
\end{figure}

\paragraph{The raw statistic reverses.}
The human study compared transitive completions after \cond{gp} and \cond{cue} primes.
In LMs this contrast is reliably positive in only 4 of 12 models and significantly negative in 7 of the 8 Llama and Qwen models, most strongly in Llama-3.1-70B, where garden-path primes lower the proportion of transitive completions by 17.6 points (Figure~\ref{fig:generation}).
After a garden-path prime, up to 29\% of completions are themselves comma-less garden-path sentences, against 0.2\% to 2.4\% after a comma prime.
These completions produce the resolved structure of the prime, a new clause after an object-less verb, in the prime's comma-less style.
They are the adaptation of \S\ref{sec:twoforces} expressed in production, and the raw contrast counts them as evidence against the residue.
Across models, the more such completions a model writes, the more negative its raw contrast ($\rho=-0.82$).
Without them, the raw contrast is positive in all 12 models (median 10 points).
Human participants completed the fragments aloud, without punctuation.
Matching primes for punctuation, as Equation~\ref{eq:A} does, is necessary to transfer the paradigm.

\paragraph{Lingering comprehension is a different measure.}
LMs answer \ex{Did the man hunt the deer?} with yes after the garden-path sentence much more often than after its comma version, and more so when the ambiguous region is longer, as human readers do \citep{christianson2001}.
Across the items of \citet{christianson2001}, however, this lingering is not positively related to the priming residue ($r$ between $-0.30$ and $0.15$ in the ten models that answer the unambiguous control questions correctly), although both measures have split-half reliabilities between 0.77 and 0.98.
Question answering and priming measure different properties of the model, so studies of lingering in LMs \citep{li2024incremental,amouyal2025} and the residue measured here are not interchangeable.

\section{Discussion and Conclusion}

A garden-path sentence leaves both analyses in an LM's context.
The abandoned one is read from the noun phrase, whose states were computed before the disambiguation, and its effect on later production grows with how strongly it was built.
The resolved one is read from the states after the disambiguation, and its effect grows with the disambiguation error.
Residual activation describes the first effect and error-driven learning the second, with in-context adaptation in the role of learning \citep{zhou2025icl}, and the two representations of the competing-representations view of lingering \citep{slattery2013} occupy different positions of the context.

The patching result was not a foregone conclusion, since later text could have relied on the main-clause states, which are more recent and encode the corrected analysis, and in the 32B and 70B models it reads almost the whole residue from the noun phrase instead.
A disambiguating word or a correcting sentence changes what later positions compute without removing the earlier representation, so appending a correction is a weaker intervention than removing the corrected text, which bears on why models refer to entities a sentence never introduced \citep{schuster2022} and rarely retract wrong answers \citep{yang2025retraction}.
Corrected statements persist in the same way, since eight of nine models still prefer a value after it has been corrected (\S\ref{sec:facts}), so applications that update what a model knows by appending a correction to its context leave the overridden version as an input to later predictions.

For human research, varying only the continuation of a fixed garden-path prefix would separate the two accounts: residual activation predicts no change in later comprehension of garden paths because the abandoned analysis is built identically, whereas error-driven learning predicts more adaptation after more surprising disambiguations, as in LMs.

\section*{Limitations}

We study English NP/Z garden paths in depth and NP/S and main-verb/reduced-relative garden paths in single experiments, so other languages and ambiguity types may behave differently.
Our models are open-weight models up to 70B parameters, because the analyses require exact log-probabilities and access to internal states, and closed models were not tested.
The behavioural manipulation of the abandoned analysis (Appendix~\ref{app:e014}) did not pass its pre-specified check, so the causal evidence for the residue comes from patching and the role-direction intervention, which were run on six to nine of the 24 models.
The contrast of Equation~\ref{eq:A} was defined after inspecting the first two models of the discovery set, and its confirmation rests on held-out items, a prospective test and pre-specified predictions, some of which failed (Appendix~\ref{app:prereg}).
The overridden statements are templated, and one pre-specified prediction about their mechanism failed, which led to the different-carrier interpretation of \S\ref{sec:facts}.
The coding of sampled completions is rule-based and was checked against gold sentences and by one annotator on a sample of generations.
We derive a prediction for human readers but do not test it.

\section*{Ethical Considerations}

This work involves no human participants and no personal data.
All models are public checkpoints used for scoring and sampling on local hardware.
The materials come from published studies and benchmarks and are used for research with citation.

\bibliography{refs}

\begin{thebibliography}{56}
\providecommand{\natexlab}[1]{#1}

\bibitem[{Amouyal et~al.(2025)Amouyal, Meltzer-Asscher, and Berant}]{amouyal2025}
Samuel~Joseph Amouyal, Aya Meltzer-Asscher, and Jonathan Berant. 2025.
\newblock \href {https://doi.org/10.18653/v1/2025.acl-long.403} {When the {LM} misunderstood the human chuckled: Analyzing garden path effects in humans and language models}.
\newblock In \emph{Proceedings of the 63rd Annual Meeting of the {A}ssociation for {C}omputational {L}inguistics (Volume 1: Long Papers)}, pages 8235--8253, Vienna, Austria. Association for Computational Linguistics.

\bibitem[{Amouyal et~al.(2026)Amouyal, Meltzer-Asscher, and Berant}]{amouyal2026}
Samuel~Joseph Amouyal, Aya Meltzer-Asscher, and Jonathan Berant. 2026.
\newblock \href {https://doi.org/10.18653/v1/2026.acl-long.1040} {Comparing human and language models sentence processing difficulties on complex structures}.
\newblock In \emph{Proceedings of the 64th Annual Meeting of the {A}ssociation for {C}omputational {L}inguistics (Volume 1: Long Papers)}, pages 22704--22722, San Diego, California, United States. Association for Computational Linguistics.

\bibitem[{Arehalli et~al.(2022)Arehalli, Dillon, and Linzen}]{arehalli2022}
Suhas Arehalli, Brian Dillon, and Tal Linzen. 2022.
\newblock \href {https://doi.org/10.18653/v1/2022.conll-1.20} {Syntactic surprisal from neural models predicts, but underestimates, human processing difficulty from syntactic ambiguities}.
\newblock In \emph{Proceedings of the 26th Conference on Computational Natural Language Learning ({C}o{NLL})}, pages 301--313, Abu Dhabi, United Arab Emirates (Hybrid). Association for Computational Linguistics.

\bibitem[{Biderman et~al.(2023)Biderman, Schoelkopf, Anthony, Bradley, O'Brien, Hallahan, Khan, Purohit, Prashanth, Raff, Skowron, Sutawika, and van~der Wal}]{biderman2023pythia}
Stella Biderman, Hailey Schoelkopf, Quentin Anthony, Herbie Bradley, Kyle O'Brien, Eric Hallahan, Mohammad~Aflah Khan, Shivanshu Purohit, USVSN~Sai Prashanth, Edward Raff, Aviya Skowron, Lintang Sutawika, and Oskar van~der Wal. 2023.
\newblock \href {https://arxiv.org/abs/2304.01373} {{Pythia}: A suite for analyzing large language models across training and scaling}.
\newblock \emph{Preprint}, arXiv:2304.01373.

\bibitem[{Bock(1986)}]{bock1986}
J.Kathryn Bock. 1986.
\newblock \href {https://doi.org/10.1016/0010-0285(86)90004-6} {Syntactic persistence in language production}.
\newblock \emph{Cognitive Psychology}, 18(3):355–387.

\bibitem[{Ceháková and Chromý(2025)}]{cehakova2025}
Markéta Ceháková and Jan Chromý. 2025.
\newblock \href {https://doi.org/10.1080/0163853x.2025.2513210} {Beyond initial misanalysis: the disruptive nature of garden-path sentences processing}.
\newblock \emph{Discourse Processes}, 62(10):723–743.

\bibitem[{Chang et~al.(2006)Chang, Dell, and Bock}]{chang2006}
Franklin Chang, Gary~S. Dell, and Kathryn Bock. 2006.
\newblock \href {https://doi.org/10.1037/0033-295x.113.2.234} {Becoming syntactic.}
\newblock \emph{Psychological Review}, 113(2):234–272.

\bibitem[{Christianson et~al.(2001)Christianson, Hollingworth, Halliwell, and Ferreira}]{christianson2001}
Kiel Christianson, Andrew Hollingworth, John~F. Halliwell, and Fernanda Ferreira. 2001.
\newblock \href {https://doi.org/10.1006/cogp.2001.0752} {Thematic roles assigned along the garden path linger}.
\newblock \emph{Cognitive Psychology}, 42(4):368–407.

\bibitem[{DerSimonian and Laird(1986)}]{dersimonian1986}
Rebecca DerSimonian and Nan Laird. 1986.
\newblock \href {https://doi.org/10.1016/0197-2456(86)90046-2} {Meta-analysis in clinical trials}.
\newblock \emph{Controlled Clinical Trials}, 7(3):177–188.

\bibitem[{Ferreira(2003)}]{ferreira2003}
F~Ferreira. 2003.
\newblock \href {https://doi.org/10.1016/s0010-0285(03)00005-7} {The misinterpretation of noncanonical sentences}.
\newblock \emph{Cognitive Psychology}, 47(2):164–203.

\bibitem[{Fine et~al.(2013)Fine, Jaeger, Farmer, and Qian}]{fine2013}
Alex~B. Fine, T.~Florian Jaeger, Thomas~A. Farmer, and Ting Qian. 2013.
\newblock \href {https://doi.org/10.1371/journal.pone.0077661} {Rapid expectation adaptation during syntactic comprehension}.
\newblock \emph{PLoS ONE}, 8(10):e77661.

\bibitem[{Frazier and Rayner(1982)}]{frazier1982}
Lyn Frazier and Keith Rayner. 1982.
\newblock \href {https://doi.org/10.1016/0010-0285(82)90008-1} {Making and correcting errors during sentence comprehension: Eye movements in the analysis of structurally ambiguous sentences}.
\newblock \emph{Cognitive Psychology}, 14(2):178–210.

\bibitem[{Fujita and Cunnings(2020)}]{fujita2020}
Hiroki Fujita and Ian Cunnings. 2020.
\newblock \href {https://doi.org/10.1017/s0142716420000351} {Lingering misinterpretation in native and nonnative sentence processing: Evidence from structural priming}.
\newblock \emph{Applied Psycholinguistics}, 42(2):475–504.

\bibitem[{Futrell et~al.(2019)Futrell, Wilcox, Morita, Qian, Ballesteros, and Levy}]{futrell2019}
Richard Futrell, Ethan Wilcox, Takashi Morita, Peng Qian, Miguel Ballesteros, and Roger Levy. 2019.
\newblock \href {https://doi.org/10.18653/v1/N19-1004} {Neural language models as psycholinguistic subjects: Representations of syntactic state}.
\newblock In \emph{Proceedings of the 2019 Conference of the {N}orth {A}merican Chapter of the {A}ssociation for {C}omputational {L}inguistics: Human Language Technologies, Volume 1 (Long and Short Papers)}, pages 32--42, Minneapolis, Minnesota. Association for Computational Linguistics.

\bibitem[{Gauthier et~al.(2020)Gauthier, Hu, Wilcox, Qian, and Levy}]{gauthier2020}
Jon Gauthier, Jennifer Hu, Ethan Wilcox, Peng Qian, and Roger Levy. 2020.
\newblock \href {https://doi.org/10.18653/v1/2020.acl-demos.10} {{S}yntax{G}ym: An online platform for targeted evaluation of language models}.
\newblock In \emph{Proceedings of the 58th Annual Meeting of the {A}ssociation for {C}omputational {L}inguistics: System Demonstrations}, pages 70--76, Online. Association for Computational Linguistics.

\bibitem[{Grattafiori et~al.(2024)Grattafiori, Dubey, Jauhri, Pandey, Kadian, Al-Dahle, Letman, Mathur, Schelten, Vaughan, Yang, Fan, Goyal, Hartshorn, Yang, Mitra, Sravankumar, Korenev, Hinsvark, Rao, Zhang, Rodriguez, Gregerson, Spataru, Roziere, Biron, Tang, Chern, Caucheteux, Nayak, Bi, Marra, McConnell, Keller, Touret, Wu, Wong, Ferrer, Nikolaidis, Allonsius, Song, Pintz, Livshits, Wyatt, Esiobu, Choudhary, Mahajan, Garcia-Olano, Perino, Hupkes, Lakomkin, AlBadawy, Lobanova, Dinan, Smith, Radenovic, Guzmán, Zhang, Synnaeve, Lee, Anderson, Thattai, Nail, Mialon, Pang, Cucurell, Nguyen, Korevaar, Xu, Touvron, Zarov, Ibarra, Kloumann, Misra, Evtimov, Zhang, Copet, Lee, Geffert, Vranes, Park, Mahadeokar, Shah, van~der Linde, Billock, Hong, Lee, Fu, Chi, Huang, Liu, Wang, Yu, Bitton, Spisak, Park, Rocca, Johnstun, Saxe, Jia, Alwala, Prasad, Upasani, Plawiak, Li, Heafield, Stone, El-Arini, Iyer, Malik, Chiu, Bhalla, Lakhotia, Rantala-Yeary, van~der Maaten, Chen, Tan, Jenkins, Martin, Madaan, Malo, Blecher, Landzaat, de~Oliveira, Muzzi, Pasupuleti, Singh, Paluri, Kardas, Tsimpoukelli, Oldham, Rita, Pavlova, Kambadur, Lewis, Si, Singh, Hassan, Goyal, Torabi, Bashlykov, Bogoychev, Chatterji, Zhang, Duchenne, Çelebi, Alrassy, Zhang, Li, Vasic, Weng, Bhargava, Dubal, Krishnan, Koura, Xu, He, Dong, Srinivasan, Ganapathy, Calderer, Cabral, Stojnic, Raileanu, Maheswari, Girdhar, Patel, Sauvestre, Polidoro, Sumbaly, Taylor, Silva, Hou, Wang, Hosseini, Chennabasappa, Singh, Bell, Kim, Edunov, Nie, Narang, Raparthy, Shen, Wan, Bhosale, Zhang, Vandenhende, Batra, Whitman, Sootla, Collot, Gururangan, Borodinsky, Herman, Fowler, Sheasha, Georgiou, Scialom, Speckbacher, Mihaylov, Xiao, Karn, Goswami, Gupta, Ramanathan, Kerkez, Gonguet, Do, Vogeti, Albiero, Petrovic, Chu, Xiong, Fu, Meers, Martinet, Wang, Wang, Tan, Xia, Xie, Jia, Wang, Goldschlag, Gaur, Babaei, Wen, Song, Zhang, Li, Mao, Coudert, Yan, Chen, Papakipos, Singh, Srivastava, Jain, Kelsey, Shajnfeld, Gangidi, Victoria, Goldstand, Menon, Sharma, Boesenberg, Baevski, Feinstein, Kallet, Sangani, Teo, Yunus, Lupu, Alvarado, Caples, Gu, Ho, Poulton, Ryan, Ramchandani, Dong, Franco, Goyal, Saraf, Chowdhury, Gabriel, Bharambe, Eisenman, Yazdan, James, Maurer, Leonhardi, Huang, Loyd, Paola, Paranjape, Liu, Wu, Ni, Hancock, Wasti, Spence, Stojkovic, Gamido, Montalvo, Parker, Burton, Mejia, Liu, Wang, Kim, Zhou, Hu, Chu, Cai, Tindal, Feichtenhofer, Gao, Civin, Beaty, Kreymer, Li, Adkins, Xu, Testuggine, David, Parikh, Liskovich, Foss, Wang, Le, Holland, Dowling, Jamil, Montgomery, Presani, Hahn, Wood, Le, Brinkman, Arcaute, Dunbar, Smothers, Sun, Kreuk, Tian, Kokkinos, Ozgenel, Caggioni, Kanayet, Seide, Florez, Schwarz, Badeer, Swee, Halpern, Herman, Sizov, Guangyi, Zhang, Lakshminarayanan, Inan, Shojanazeri, Zou, Wang, Zha, Habeeb, Rudolph, Suk, Aspegren, Goldman, Zhan, Damlaj, Molybog, Tufanov, Leontiadis, Veliche, Gat, Weissman, Geboski, Kohli, Lam, Asher, Gaya, Marcus, Tang, Chan, Zhen, Reizenstein, Teboul, Zhong, Jin, Yang, Cummings, Carvill, Shepard, McPhie, Torres, Ginsburg, Wang, Wu, U, Saxena, Khandelwal, Zand, Matosich, Veeraraghavan, Michelena, Li, Jagadeesh, Huang, Chawla, Huang, Chen, Garg, A, Silva, Bell, Zhang, Guo, Yu, Moshkovich, Wehrstedt, Khabsa, Avalani, Bhatt, Mankus, Hasson, Lennie, Reso, Groshev, Naumov, Lathi, Keneally, Liu, Seltzer, Valko, Restrepo, Patel, Vyatskov, Samvelyan, Clark, Macey, Wang, Hermoso, Metanat, Rastegari, Bansal, Santhanam, Parks, White, Bawa, Singhal, Egebo, Usunier, Mehta, Laptev, Dong, Cheng, Chernoguz, Hart, Salpekar, Kalinli, Kent, Parekh, Saab, Balaji, Rittner, Bontrager, Roux, Dollar, Zvyagina, Ratanchandani, Yuvraj, Liang, Alao, Rodriguez, Ayub, Murthy, Nayani, Mitra, Parthasarathy, Li, Hogan, Battey, Wang, Howes, Rinott, Mehta, Siby, Bondu, Datta, Chugh, Hunt, Dhillon, Sidorov, Pan, Mahajan, Verma, Yamamoto, Ramaswamy, Lindsay, Lindsay, Feng, Lin, Zha, Patil, Shankar, Zhang, Zhang, Wang, Agarwal, Sajuyigbe, Chintala, Max, Chen, Kehoe, Satterfield, Govindaprasad, Gupta, Deng, Cho, Virk, Subramanian, Choudhury, Goldman, Remez, Glaser, Best, Koehler, Robinson, Li, Zhang, Matthews, Chou, Shaked, Vontimitta, Ajayi, Montanez, Mohan, Kumar, Mangla, Ionescu, Poenaru, Mihailescu, Ivanov, Li, Wang, Jiang, Bouaziz, Constable, Tang, Wu, Wang, Wu, Gao, Kleinman, Chen, Hu, Jia, Qi, Li, Zhang, Zhang, Adi, Nam, Yu, Wang, Zhao, Hao, Qian, Li, He, Rait, DeVito, Rosnbrick, Wen, Yang, Zhao, and Ma}]{grattafiori2024llama}
Aaron Grattafiori, Abhimanyu Dubey, Abhinav Jauhri, Abhinav Pandey, Abhishek Kadian, Ahmad Al-Dahle, Aiesha Letman, Akhil Mathur, Alan Schelten, Alex Vaughan, Amy Yang, Angela Fan, Anirudh Goyal, Anthony Hartshorn, Aobo Yang, Archi Mitra, Archie Sravankumar, Artem Korenev, Arthur Hinsvark, and 542 others. 2024.
\newblock \href {https://arxiv.org/abs/2407.21783} {The {Llama} 3 herd of models}.
\newblock \emph{Preprint}, arXiv:2407.21783.

\bibitem[{Hanna and Mueller(2025)}]{hanna2025}
Michael Hanna and Aaron Mueller. 2025.
\newblock \href {https://doi.org/10.18653/v1/2025.naacl-long.164} {Incremental sentence processing mechanisms in autoregressive {Transformer} language models}.
\newblock In \emph{Proceedings of the 2025 Conference of the Nations of the {A}mericas Chapter of the {A}ssociation for {C}omputational {L}inguistics: Human Language Technologies (Volume 1: Long Papers)}, pages 3181--3203, Albuquerque, New Mexico. Association for Computational Linguistics.

\bibitem[{Harrington~Stack et~al.(2018)Harrington~Stack, James, and Watson}]{harringtonstack2018}
Caoimhe~M. Harrington~Stack, Ariel~N. James, and Duane~G. Watson. 2018.
\newblock \href {https://doi.org/10.3758/s13421-018-0808-6} {A failure to replicate rapid syntactic adaptation in comprehension}.
\newblock \emph{Memory \& Cognition}, 46(6):864–877.

\bibitem[{Hartsuiker et~al.(2008)Hartsuiker, Bernolet, Schoonbaert, Speybroeck, and Vanderelst}]{hartsuiker2008}
Robert~J. Hartsuiker, Sarah Bernolet, Sofie Schoonbaert, Sara Speybroeck, and Dieter Vanderelst. 2008.
\newblock \href {https://doi.org/10.1016/j.jml.2007.07.003} {Syntactic priming persists while the lexical boost decays: Evidence from written and spoken dialogue}.
\newblock \emph{Journal of Memory and Language}, 58(2):214–238.

\bibitem[{Huang et~al.(2024)Huang, Arehalli, Kugemoto, Muxica, Prasad, Dillon, and Linzen}]{huang2024}
Kuan-Jung Huang, Suhas Arehalli, Mari Kugemoto, Christian Muxica, Grusha Prasad, Brian Dillon, and Tal Linzen. 2024.
\newblock \href {https://doi.org/10.1016/j.jml.2024.104510} {Large-scale benchmark yields no evidence that language model surprisal explains syntactic disambiguation difficulty}.
\newblock \emph{Journal of Memory and Language}, 137:104510.

\bibitem[{Jaeger and Snider(2013)}]{jaeger2013}
T.~Florian Jaeger and Neal~E. Snider. 2013.
\newblock \href {https://doi.org/10.1016/j.cognition.2012.10.013} {Alignment as a consequence of expectation adaptation: Syntactic priming is affected by the prime’s prediction error given both prior and recent experience}.
\newblock \emph{Cognition}, 127(1):57–83.

\bibitem[{Jiang et~al.(2023)Jiang, Sablayrolles, Mensch, Bamford, Chaplot, de~las Casas, Bressand, Lengyel, Lample, Saulnier, Lavaud, Lachaux, Stock, Scao, Lavril, Wang, Lacroix, and Sayed}]{jiang2023mistral}
Albert~Q. Jiang, Alexandre Sablayrolles, Arthur Mensch, Chris Bamford, Devendra~Singh Chaplot, Diego de~las Casas, Florian Bressand, Gianna Lengyel, Guillaume Lample, Lucile Saulnier, Lélio~Renard Lavaud, Marie-Anne Lachaux, Pierre Stock, Teven~Le Scao, Thibaut Lavril, Thomas Wang, Timothée Lacroix, and William~El Sayed. 2023.
\newblock \href {https://arxiv.org/abs/2310.06825} {{Mistral} {7B}}.
\newblock \emph{Preprint}, arXiv:2310.06825.

\bibitem[{Jumelet et~al.(2024)Jumelet, Zuidema, and Sinclair}]{jumelet2024}
Jaap Jumelet, Willem Zuidema, and Arabella Sinclair. 2024.
\newblock \href {https://doi.org/10.18653/v1/2024.findings-acl.877} {Do language models exhibit human-like structural priming effects?}
\newblock In \emph{Findings of the {A}ssociation for {C}omputational {L}inguistics: {ACL} 2024}, pages 14727--14742, Bangkok, Thailand. Association for Computational Linguistics.

\bibitem[{Kamoi et~al.(2024)Kamoi, Zhang, Zhang, Han, and Zhang}]{kamoi2024}
Ryo Kamoi, Yusen Zhang, Nan Zhang, Jiawei Han, and Rui Zhang. 2024.
\newblock \href {https://doi.org/10.1162/tacl_a_00713} {When can {LLM}s actually correct their own mistakes? a critical survey of self-correction of {LLM}s}.
\newblock \emph{Transactions of the Association for Computational Linguistics}, 12:1417--1440.

\bibitem[{Kwon et~al.(2023)Kwon, Li, Zhuang, Sheng, Zheng, Yu, Gonzalez, Zhang, and Stoica}]{kwon2023vllm}
Woosuk Kwon, Zhuohan Li, Siyuan Zhuang, Ying Sheng, Lianmin Zheng, Cody~Hao Yu, Joseph Gonzalez, Hao Zhang, and Ion Stoica. 2023.
\newblock \href {https://doi.org/10.1145/3600006.3613165} {Efficient memory management for large language model serving with {PagedAttention}}.
\newblock In \emph{Proceedings of the 29th Symposium on Operating Systems Principles}, SOSP ’23, page 611–626. ACM.

\bibitem[{Li et~al.(2024)Li, Feng, Narang, Peng, Cai, Shah, and Varma}]{li2024incremental}
Andrew Li, Xianle Feng, Siddhant Narang, Austin Peng, Tianle Cai, Raj~Sanjay Shah, and Sashank Varma. 2024.
\newblock \href {https://arxiv.org/abs/2405.16042} {Incremental comprehension of garden-path sentences by large language models: Semantic interpretation, syntactic re-analysis, and attention}.
\newblock \emph{Preprint}, arXiv:2405.16042.

\bibitem[{Longpre et~al.(2021)Longpre, Perisetla, Chen, Ramesh, DuBois, and Singh}]{longpre2021}
Shayne Longpre, Kartik Perisetla, Anthony Chen, Nikhil Ramesh, Chris DuBois, and Sameer Singh. 2021.
\newblock \href {https://doi.org/10.18653/v1/2021.emnlp-main.565} {Entity-based knowledge conflicts in question answering}.
\newblock In \emph{Proceedings of the 2021 Conference on Empirical Methods in Natural Language Processing}, pages 7052--7063, Online and Punta Cana, Dominican Republic. Association for Computational Linguistics.

\bibitem[{Meng et~al.(2023)Meng, Bau, Andonian, and Belinkov}]{meng2022locating}
Kevin Meng, David Bau, Alex Andonian, and Yonatan Belinkov. 2023.
\newblock \href {https://arxiv.org/abs/2202.05262} {Locating and editing factual associations in {GPT}}.
\newblock \emph{Preprint}, arXiv:2202.05262.

\bibitem[{Michaelov et~al.(2023)Michaelov, Arnett, Chang, and Bergen}]{michaelov2023}
James~A. Michaelov, Catherine Arnett, Tyler~A. Chang, and Benjamin~K. Bergen. 2023.
\newblock \href {https://doi.org/10.18653/v1/2023.emnlp-main.227} {Structural priming demonstrates abstract grammatical representations in multilingual language models}.
\newblock In \emph{Proceedings of the 2023 Conference on Empirical Methods in Natural Language Processing}, pages 3703--3720, Singapore. Association for Computational Linguistics.

\bibitem[{Pickering and Branigan(1998)}]{pickering1998}
Martin~J. Pickering and Holly~P. Branigan. 1998.
\newblock \href {https://doi.org/10.1006/jmla.1998.2592} {The representation of verbs: Evidence from syntactic priming in language production}.
\newblock \emph{Journal of Memory and Language}, 39(4):633–651.

\bibitem[{Prakash et~al.(2026)Prakash, Shapira, Sharma, Riedl, Belinkov, Shaham, Bau, and Geiger}]{prakash2026}
Nikhil Prakash, Natalie Shapira, Arnab~Sen Sharma, Christoph Riedl, Yonatan Belinkov, Tamar~Rott Shaham, David Bau, and Atticus Geiger. 2026.
\newblock \href {https://arxiv.org/abs/2505.14685} {Language models use lookbacks to track beliefs}.
\newblock \emph{Preprint}, arXiv:2505.14685.

\bibitem[{Prasad and Linzen(2021)}]{prasad2021}
Grusha Prasad and Tal Linzen. 2021.
\newblock \href {https://doi.org/10.1037/xlm0001046} {Rapid syntactic adaptation in self-paced reading: Detectable, but only with many participants.}
\newblock \emph{Journal of Experimental Psychology: Learning, Memory, and Cognition}, 47(7):1156–1172.

\bibitem[{Prasad et~al.(2019)Prasad, van Schijndel, and Linzen}]{prasad2019}
Grusha Prasad, Marten van Schijndel, and Tal Linzen. 2019.
\newblock \href {https://doi.org/10.18653/v1/K19-1007} {Using priming to uncover the organization of syntactic representations in neural language models}.
\newblock In \emph{Proceedings of the 23rd Conference on Computational Natural Language Learning ({C}o{NLL})}, pages 66--76, Hong Kong, China. Association for Computational Linguistics.

\bibitem[{Pucci et~al.(2026)Pucci, Li, and Sinclair}]{pucci2026}
Giulia Pucci, Ruizhe Li, and Arabella Sinclair. 2026.
\newblock \href {https://arxiv.org/abs/2609.04484} {Patterns of priming in production: Lexical, semantic and structural alignment in language model generation}.
\newblock \emph{Preprint}, arXiv:2609.04484.

\bibitem[{Qwen et~al.(2025)Qwen, :, Yang, Yang, Zhang, Hui, Zheng, Yu, Li, Liu, Huang, Wei, Lin, Yang, Tu, Zhang, Yang, Yang, Zhou, Lin, Dang, Lu, Bao, Yang, Yu, Li, Xue, Zhang, Zhu, Men, Lin, Li, Tang, Xia, Ren, Ren, Fan, Su, Zhang, Wan, Liu, Cui, Zhang, and Qiu}]{qwen2024qwen25}
Qwen, :, An~Yang, Baosong Yang, Beichen Zhang, Binyuan Hui, Bo~Zheng, Bowen Yu, Chengyuan Li, Dayiheng Liu, Fei Huang, Haoran Wei, Huan Lin, Jian Yang, Jianhong Tu, Jianwei Zhang, Jianxin Yang, Jiaxi Yang, Jingren Zhou, and 25 others. 2025.
\newblock \href {https://arxiv.org/abs/2412.15115} {{Qwen2.5} technical report}.
\newblock \emph{Preprint}, arXiv:2412.15115.

\bibitem[{Radford et~al.(2019)Radford, Wu, Child, Luan, Amodei, and Sutskever}]{radford2019language}
Alec Radford, Jeffrey Wu, Rewon Child, David Luan, Dario Amodei, and Ilya Sutskever. 2019.
\newblock Language models are unsupervised multitask learners.
\newblock \emph{OpenAI blog}.

\bibitem[{Reitter et~al.(2011)Reitter, Keller, and Moore}]{reitter2011}
David Reitter, Frank Keller, and Johanna~D. Moore. 2011.
\newblock \href {https://doi.org/10.1111/j.1551-6709.2010.01165.x} {A computational cognitive model of syntactic priming}.
\newblock \emph{Cognitive Science}, 35(4):587–637.

\bibitem[{Schuster and Linzen(2022)}]{schuster2022}
Sebastian Schuster and Tal Linzen. 2022.
\newblock \href {https://doi.org/10.18653/v1/2022.naacl-main.71} {When a sentence does not introduce a discourse entity, transformer-based models still sometimes refer to it}.
\newblock In \emph{Proceedings of the 2022 Conference of the {N}orth {A}merican Chapter of the {A}ssociation for {C}omputational {L}inguistics: Human Language Technologies}, pages 969--982, Seattle, United States. Association for Computational Linguistics.

\bibitem[{Sinclair et~al.(2022)Sinclair, Jumelet, Zuidema, and Fern{\'a}ndez}]{sinclair2022}
Arabella Sinclair, Jaap Jumelet, Willem Zuidema, and Raquel Fern{\'a}ndez. 2022.
\newblock \href {https://doi.org/10.1162/tacl_a_00504} {Structural persistence in language models: Priming as a window into abstract language representations}.
\newblock \emph{Transactions of the Association for Computational Linguistics}, 10:1031--1050.

\bibitem[{Slattery et~al.(2013)Slattery, Sturt, Christianson, Yoshida, and Ferreira}]{slattery2013}
Timothy~J. Slattery, Patrick Sturt, Kiel Christianson, Masaya Yoshida, and Fernanda Ferreira. 2013.
\newblock \href {https://doi.org/10.1016/j.jml.2013.04.001} {Lingering misinterpretations of garden path sentences arise from competing syntactic representations}.
\newblock \emph{Journal of Memory and Language}, 69(2):104–120.

\bibitem[{Sturt and Crocker(1996)}]{sturt1996}
Patrick Sturt and Matthew~W. Crocker. 1996.
\newblock \href {https://doi.org/10.1080/016909696387123} {Monotonic syntactic processing: A cross-linguistic study of attachment and reanalysis}.
\newblock \emph{Language and Cognitive Processes}, 11(5):449–494.

\bibitem[{Tabor et~al.(2004)Tabor, Galantucci, and Richardson}]{tabor2004}
Whitney Tabor, Bruno Galantucci, and Daniel Richardson. 2004.
\newblock \href {https://doi.org/10.1016/j.jml.2004.01.001} {Effects of merely local syntactic coherence on sentence processing}.
\newblock \emph{Journal of Memory and Language}, 50(4):355–370.

\bibitem[{Team et~al.(2025)Team, Kamath, Ferret, Pathak, Vieillard, Merhej, Perrin, Matejovicova, Ramé, Rivière, Rouillard, Mesnard, Cideron, bastien Grill, Ramos, Yvinec, Casbon, Pot, Penchev, Liu, Visin, Kenealy, Beyer, Zhai, Tsitsulin, Busa-Fekete, Feng, Sachdeva, Coleman, Gao, Mustafa, Barr, Parisotto, Tian, Eyal, Cherry, Peter, Sinopalnikov, Bhupatiraju, Agarwal, Kazemi, Malkin, Kumar, Vilar, Brusilovsky, Luo, Steiner, Friesen, Sharma, Sharma, Gilady, Goedeckemeyer, Saade, Feng, Kolesnikov, Bendebury, Abdagic, Vadi, György, Pinto, Das, Bapna, Miech, Yang, Paterson, Shenoy, Chakrabarti, Piot, Wu, Shahriari, Petrini, Chen, Lan, Choquette-Choo, Carey, Brick, Deutsch, Eisenbud, Cattle, Cheng, Paparas, Sreepathihalli, Reid, Tran, Zelle, Noland, Huizenga, Kharitonov, Liu, Amirkhanyan, Cameron, Hashemi, Klimczak-Plucińska, Singh, Mehta, Lehri, Hazimeh, Ballantyne, Szpektor, Nardini, Pouget-Abadie, Chan, Stanton, Wieting, Lai, Orbay, Fernandez, Newlan, yeong Ji, Singh, Black, Yu, Hui, Vodrahalli, Greff, Qiu, Valentine, Coelho, Ritter, Hoffman, Watson, Chaturvedi, Moynihan, Ma, Babar, Noy, Byrd, Roy, Momchev, Chauhan, Sachdeva, Bunyan, Botarda, Caron, Rubenstein, Culliton, Schmid, Sessa, Xu, Stanczyk, Tafti, Shivanna, Wu, Pan, Rokni, Willoughby, Vallu, Mullins, Jerome, Smoot, Girgin, Iqbal, Reddy, Sheth, Põder, Bhatnagar, Panyam, Eiger, Zhang, Liu, Yacovone, Liechty, Kalra, Evci, Misra, Roseberry, Feinberg, Kolesnikov, Han, Kwon, Chen, Chow, Zhu, Wei, Egyed, Cotruta, Giang, Kirk, Rao, Black, Babar, Lo, Moreira, Martins, Sanseviero, Gonzalez, Gleicher, Warkentin, Mirrokni, Senter, Collins, Barral, Ghahramani, Hadsell, Matias, Sculley, Petrov, Fiedel, Shazeer, Vinyals, Dean, Hassabis, Kavukcuoglu, Farabet, Buchatskaya, Alayrac, Anil, Dmitry, Lepikhin, Borgeaud, Bachem, Joulin, Andreev, Hardin, Dadashi, and Hussenot}]{gemma2025gemma3}
Gemma Team, Aishwarya Kamath, Johan Ferret, Shreya Pathak, Nino Vieillard, Ramona Merhej, Sarah Perrin, Tatiana Matejovicova, Alexandre Ramé, Morgane Rivière, Louis Rouillard, Thomas Mesnard, Geoffrey Cideron, Jean bastien Grill, Sabela Ramos, Edouard Yvinec, Michelle Casbon, Etienne Pot, Ivo Penchev, and 197 others. 2025.
\newblock \href {https://arxiv.org/abs/2503.19786} {{Gemma} 3 technical report}.
\newblock \emph{Preprint}, arXiv:2503.19786.

\bibitem[{Team et~al.(2024)Team, Riviere, Pathak, Sessa, Hardin, Bhupatiraju, Hussenot, Mesnard, Shahriari, Ramé, Ferret, Liu, Tafti, Friesen, Casbon, Ramos, Kumar, Lan, Jerome, Tsitsulin, Vieillard, Stanczyk, Girgin, Momchev, Hoffman, Thakoor, Grill, Neyshabur, Bachem, Walton, Severyn, Parrish, Ahmad, Hutchison, Abdagic, Carl, Shen, Brock, Coenen, Laforge, Paterson, Bastian, Piot, Wu, Royal, Chen, Kumar, Perry, Welty, Choquette-Choo, Sinopalnikov, Weinberger, Vijaykumar, Rogozińska, Herbison, Bandy, Wang, Noland, Moreira, Senter, Eltyshev, Visin, Rasskin, Wei, Cameron, Martins, Hashemi, Klimczak-Plucińska, Batra, Dhand, Nardini, Mein, Zhou, Svensson, Stanway, Chan, Zhou, Carrasqueira, Iljazi, Becker, Fernandez, van Amersfoort, Gordon, Lipschultz, Newlan, yeong Ji, Mohamed, Badola, Black, Millican, McDonell, Nguyen, Sodhia, Greene, Sjoesund, Usui, Sifre, Heuermann, Lago, McNealus, Soares, Kilpatrick, Dixon, Martins, Reid, Singh, Iverson, Görner, Velloso, Wirth, Davidow, Miller, Rahtz, Watson, Risdal, Kazemi, Moynihan, Zhang, Kahng, Park, Rahman, Khatwani, Dao, Bardoliwalla, Devanathan, Dumai, Chauhan, Wahltinez, Botarda, Barnes, Barham, Michel, Jin, Georgiev, Culliton, Kuppala, Comanescu, Merhej, Jana, Rokni, Agarwal, Mullins, Saadat, Carthy, Cogan, Perrin, Arnold, Krause, Dai, Garg, Sheth, Ronstrom, Chan, Jordan, Yu, Eccles, Hennigan, Kocisky, Doshi, Jain, Yadav, Meshram, Dharmadhikari, Barkley, Wei, Ye, Han, Kwon, Xu, Shen, Gong, Wei, Cotruta, Kirk, Rao, Giang, Peran, Warkentin, Collins, Barral, Ghahramani, Hadsell, Sculley, Banks, Dragan, Petrov, Vinyals, Dean, Hassabis, Kavukcuoglu, Farabet, Buchatskaya, Borgeaud, Fiedel, Joulin, Kenealy, Dadashi, and Andreev}]{gemma2024gemma2}
Gemma Team, Morgane Riviere, Shreya Pathak, Pier~Giuseppe Sessa, Cassidy Hardin, Surya Bhupatiraju, Léonard Hussenot, Thomas Mesnard, Bobak Shahriari, Alexandre Ramé, Johan Ferret, Peter Liu, Pouya Tafti, Abe Friesen, Michelle Casbon, Sabela Ramos, Ravin Kumar, Charline~Le Lan, Sammy Jerome, and 179 others. 2024.
\newblock \href {https://arxiv.org/abs/2408.00118} {{Gemma} 2: Improving open language models at a practical size}.
\newblock \emph{Preprint}, arXiv:2408.00118.

\bibitem[{Timkey et~al.(2026)Timkey, Dillon, and Linzen}]{timkey2026}
William Timkey, Brian Dillon, and Tal Linzen. 2026.
\newblock \href {https://arxiv.org/abs/2605.15440} {Why are language models less surprised than humans? testing the parse multiplicity mismatch hypothesis}.
\newblock \emph{Preprint}, arXiv:2605.15440.

\bibitem[{Touvron et~al.(2023)Touvron, Martin, Stone, Albert, Almahairi, Babaei, Bashlykov, Batra, Bhargava, Bhosale, Bikel, Blecher, Ferrer, Chen, Cucurull, Esiobu, Fernandes, Fu, Fu, Fuller, Gao, Goswami, Goyal, Hartshorn, Hosseini, Hou, Inan, Kardas, Kerkez, Khabsa, Kloumann, Korenev, Koura, Lachaux, Lavril, Lee, Liskovich, Lu, Mao, Martinet, Mihaylov, Mishra, Molybog, Nie, Poulton, Reizenstein, Rungta, Saladi, Schelten, Silva, Smith, Subramanian, Tan, Tang, Taylor, Williams, Kuan, Xu, Yan, Zarov, Zhang, Fan, Kambadur, Narang, Rodriguez, Stojnic, Edunov, and Scialom}]{touvron2023llama}
Hugo Touvron, Louis Martin, Kevin Stone, Peter Albert, Amjad Almahairi, Yasmine Babaei, Nikolay Bashlykov, Soumya Batra, Prajjwal Bhargava, Shruti Bhosale, Dan Bikel, Lukas Blecher, Cristian~Canton Ferrer, Moya Chen, Guillem Cucurull, David Esiobu, Jude Fernandes, Jeremy Fu, Wenyin Fu, and 49 others. 2023.
\newblock \href {https://arxiv.org/abs/2307.09288} {{Llama} 2: Open foundation and fine-tuned chat models}.
\newblock \emph{Preprint}, arXiv:2307.09288.

\bibitem[{van Gompel et~al.(2006)van Gompel, Pickering, Pearson, and Jacob}]{vangompel2006}
Roger~P.G. van Gompel, Martin~J. Pickering, Jamie Pearson, and Gunnar Jacob. 2006.
\newblock \href {https://doi.org/10.1016/j.jml.2006.06.004} {The activation of inappropriate analyses in garden-path sentences: Evidence from structural priming}.
\newblock \emph{Journal of Memory and Language}, 55(3):335–362.

\bibitem[{van Schijndel and Linzen(2018)}]{vanschijndel2018}
Marten van Schijndel and Tal Linzen. 2018.
\newblock \href {https://doi.org/10.18653/v1/D18-1499} {A neural model of adaptation in reading}.
\newblock In \emph{Proceedings of the 2018 Conference on Empirical Methods in Natural Language Processing}, pages 4704--4710, Brussels, Belgium. Association for Computational Linguistics.

\bibitem[{van Schijndel and Linzen(2021)}]{vanschijndel2021}
Marten van Schijndel and Tal Linzen. 2021.
\newblock \href {https://doi.org/10.1111/cogs.12988} {Single‐stage prediction models do not explain the magnitude of syntactic disambiguation difficulty}.
\newblock \emph{Cognitive Science}, 45(6).

\bibitem[{Vig et~al.(2020)Vig, Gehrmann, Belinkov, Qian, Nevo, Sakenis, Huang, Singer, and Shieber}]{vig2020causal}
Jesse Vig, Sebastian Gehrmann, Yonatan Belinkov, Sharon Qian, Daniel Nevo, Simas Sakenis, Jason Huang, Yaron Singer, and Stuart Shieber. 2020.
\newblock \href {https://arxiv.org/abs/2004.12265} {Causal mediation analysis for interpreting neural {NLP}: The case of gender bias}.
\newblock \emph{Preprint}, arXiv:2004.12265.

\bibitem[{Xie et~al.(2024)Xie, Zhang, Chen, Lou, and Su}]{xie2024conflict}
Jian Xie, Kai Zhang, Jiangjie Chen, Renze Lou, and Yu~Su. 2024.
\newblock \href {https://arxiv.org/abs/2305.13300} {Adaptive chameleon or stubborn sloth: Revealing the behavior of large language models in knowledge conflicts}.
\newblock \emph{Preprint}, arXiv:2305.13300.

\bibitem[{Yang et~al.(2025)Yang, Li, Yang, Zhang, Hui, Zheng, Yu, Gao, Huang, Lv, Zheng, Liu, Zhou, Huang, Hu, Ge, Wei, Lin, Tang, Yang, Tu, Zhang, Yang, Yang, Zhou, Zhou, Lin, Dang, Bao, Yang, Yu, Deng, Li, Xue, Li, Zhang, Wang, Zhu, Men, Gao, Liu, Luo, Li, Tang, Yin, Ren, Wang, Zhang, Ren, Fan, Su, Zhang, Zhang, Wan, Liu, Wang, Cui, Zhang, Zhou, and Qiu}]{yang2025qwen3}
An~Yang, Anfeng Li, Baosong Yang, Beichen Zhang, Binyuan Hui, Bo~Zheng, Bowen Yu, Chang Gao, Chengen Huang, Chenxu Lv, Chujie Zheng, Dayiheng Liu, Fan Zhou, Fei Huang, Feng Hu, Hao Ge, Haoran Wei, Huan Lin, Jialong Tang, and 41 others. 2025.
\newblock \href {https://arxiv.org/abs/2505.09388} {{Qwen3} technical report}.
\newblock \emph{Preprint}, arXiv:2505.09388.

\bibitem[{Yang and Jia(2026)}]{yang2025retraction}
Yuqing Yang and Robin Jia. 2026.
\newblock \href {https://arxiv.org/abs/2505.16170} {When do {LLMs} admit their mistakes? understanding the role of model belief in retraction}.
\newblock \emph{Preprint}, arXiv:2505.16170.

\bibitem[{Yoshida et~al.(2026)Yoshida, Isono, Someya, Oseki, and Kuribayashi}]{yoshida2026}
Ryo Yoshida, Shinnosuke Isono, Taiga Someya, Yohei Oseki, and Tatsuki Kuribayashi. 2026.
\newblock \href {https://doi.org/10.18653/v1/2026.acl-long.1694} {An existence proof for neural language models that can explain garden-path effects via surprisal}.
\newblock In \emph{Proceedings of the 64th Annual Meeting of the {A}ssociation for {C}omputational {L}inguistics (Volume 1: Long Papers)}, pages 36556--36569, San Diego, California, United States. Association for Computational Linguistics.

\bibitem[{Zhang and Nanda(2024)}]{zhang2024patching}
Fred Zhang and Neel Nanda. 2024.
\newblock \href {https://arxiv.org/abs/2309.16042} {Towards best practices of activation patching in language models: Metrics and methods}.
\newblock \emph{Preprint}, arXiv:2309.16042.

\bibitem[{Zhou et~al.(2025)Zhou, Frank, and McCoy}]{zhou2025icl}
Zhenghao Zhou, Robert Frank, and R.~Thomas McCoy. 2025.
\newblock \href {https://doi.org/10.18653/v1/2025.naacl-long.586} {Is in-context learning a type of error-driven learning? evidence from the inverse frequency effect in structural priming}.
\newblock In \emph{Proceedings of the 2025 Conference of the Nations of the {A}mericas Chapter of the {A}ssociation for {C}omputational {L}inguistics: Human Language Technologies (Volume 1: Long Papers)}, pages 11712--11725, Albuquerque, New Mexico. Association for Computational Linguistics.

\end{thebibliography}

\appendix

\section{Models}
\label{app:models}

Table~\ref{tab:models} lists the 24 models.
All weights are public Hugging Face checkpoints: GPT-2 \citep{radford2019language}, Pythia \citep{biderman2023pythia}, Llama~2 and Llama~3 \citep{touvron2023llama,grattafiori2024llama}, Mistral \citep{jiang2023mistral}, Gemma~2 and Gemma~3 \citep{gemma2024gemma2,gemma2025gemma3}, Qwen2.5 \citep{qwen2024qwen25} and Qwen3 \citep{yang2025qwen3}.
Pythia-410M is the seed-1 checkpoint of the Pythia suite.
Log-probabilities are computed on the jointly tokenized prime and target, with the prime preceded by the beginning-of-sequence token and separated from the target by a space.
A target token counts toward the continuation if its character span starts at or after the continuation's first character.
Priming scores were computed in bfloat16.
Patching runs used float32 for models up to 8B parameters, where the cached and the full-sequence computations agree to $10^{-4}$ nats, and bfloat16 for Qwen2.5-32B and Llama-3.1-70B, which ran on two GPUs.
Sampled completions were generated with vLLM \citep{kwon2023vllm} using temperature 1, top-$p$ 1, at most 24 new tokens and a stop at the first line break.

\begin{table*}[t]
\centering\footnotesize
\setlength{\tabcolsep}{3pt}
\begin{adjustbox}{max width=\textwidth}
\begin{tabular}{lllcccccccc}
\toprule
Model & Size & Type & Disc. & Held-out & \cond{nom} & \cond{pass} & Christianson & Sampled & Patching & Context\\
\midrule
GPT-2 & 0.12B & base & \checkmark & \checkmark & \checkmark & \checkmark & \checkmark &  &  & \\
Pythia-160M & 0.16B & base & \checkmark & \checkmark & \checkmark & \checkmark & \checkmark &  &  & \\
Pythia-410M & 0.41B & base & \checkmark & \checkmark & \checkmark & \checkmark & \checkmark &  &  & \\
Pythia-1.4B & 1.4B & base & \checkmark & \checkmark & \checkmark & \checkmark & \checkmark & \checkmark & \checkmark & \checkmark\\
Pythia-2.8B & 2.8B & base & \checkmark & \checkmark & \checkmark & \checkmark & \checkmark &  & \checkmark & \checkmark\\
Qwen2.5-0.5B & 0.5B & base & \checkmark & \checkmark & \checkmark & \checkmark & \checkmark &  &  & \\
Qwen3-1.7B-Base & 1.7B & base & \checkmark & \checkmark & \checkmark & \checkmark & \checkmark &  &  & \\
Gemma-2-2B & 2.6B & base & \checkmark & \checkmark & \checkmark & \checkmark & \checkmark & \checkmark &  & \checkmark\\
Qwen2.5-3B & 3B & base & \checkmark & \checkmark & \checkmark & \checkmark & \checkmark &  &  & \\
Llama-3.2-3B & 3.2B & base & \checkmark & \checkmark & \checkmark & \checkmark & \checkmark & \checkmark & \checkmark & \checkmark\\
Llama-2-7B & 7B & base & \checkmark & \checkmark & \checkmark & \checkmark & \checkmark &  &  & \\
Mistral-7B-v0.3 & 7B & base & \checkmark & \checkmark & \checkmark & \checkmark & \checkmark & \checkmark & \checkmark & \checkmark\\
Qwen2.5-7B & 7.6B & base & \checkmark & \checkmark & \checkmark & \checkmark & \checkmark & \checkmark & \checkmark & \checkmark\\
Qwen2.5-7B-Instruct & 7.6B & instruct & \checkmark & \checkmark & \checkmark & \checkmark & \checkmark & \checkmark &  & \checkmark\\
Llama-3.1-8B-Instruct & 8B & instruct & \checkmark & \checkmark & \checkmark & \checkmark & \checkmark & \checkmark & \checkmark & \checkmark\\
Qwen3-8B-Base & 8.2B & base & \checkmark & \checkmark & \checkmark & \checkmark & \checkmark & \checkmark & \checkmark & \checkmark\\
Qwen3-8B & 8.2B & post-trained &  &  &  &  & \checkmark &  &  & \\
Gemma-2-9B & 9.2B & base & \checkmark & \checkmark & \checkmark & \checkmark & \checkmark & \checkmark &  & \checkmark\\
Gemma-3-12B-it & 12B & instruct & \checkmark & \checkmark & \checkmark & \checkmark & \checkmark &  &  & \\
Qwen2.5-14B-Instruct & 14.7B & instruct &  &  &  &  & \checkmark &  &  & \\
Qwen3-30B-A3B & 30B (3B active) & post-trained MoE & \checkmark & \checkmark & \checkmark & \checkmark & \checkmark &  &  & \\
Qwen2.5-32B & 32B & base & \checkmark & \checkmark & \checkmark & \checkmark & \checkmark & \checkmark & \checkmark & \\
Qwen3-32B & 32B & post-trained & \checkmark & \checkmark & \checkmark & \checkmark & \checkmark & \checkmark &  & \\
Llama-3.1-70B & 70B & base & \checkmark & \checkmark & \checkmark & \checkmark & \checkmark & \checkmark & \checkmark & \checkmark\\
 
\bottomrule
\end{tabular}
\end{adjustbox}
\caption{The 24 models and the experiments they enter. Disc.: discovery set (\S\ref{sec:residue}, \S\ref{sec:twoforces}). Held-out: 48 held-out families. \cond{nom}: case disambiguation. \cond{pass}: passive primes. Christianson: items of \citet{christianson2001}, used for the prospective test of \S\ref{sec:twoforces} and for comprehension questions (\S\ref{sec:transfer}). Sampled: sampled completions. Patching: \S\ref{sec:mechanism}. Context: Appendix~\ref{app:e014}.}
\label{tab:models}
\end{table*}

\section{Materials}
\label{app:materials}

The discovery families come from the released items of \citet{amouyal2025} and from the NP/Z subset of the SAP benchmark \citep{huang2024}.
For each family we wrote the comma version (\cond{cue}), the transitive versions with a resumptive pronoun (\cond{tb}, \cond{tc}), and the two intransitive versions (\cond{nb}, \cond{nc}).
The intransitive verbs (\ex{slept, yawned, sneezed, laughed, sighed, snored, napped, shivered, dozed, frowned, giggled, rested}) cannot take an object, and each family uses one of them with the family's own subject.
The held-out families combine 24 items with reflexive-absolute verbs from \citet{cehakova2025} with 24 NP/Z items from SyntaxGym \citep{gauthier2020}.
The two sources differ in the object expectation of their verbs (about $-0.3$ and $1.5$ nats on average), which is why we report between-family slopes on the held-out set with source fixed effects.
The items of \citet{christianson2001} were transcribed from the published appendix.
They comprise 42 optionally transitive and 12 reflexive-absolute verbs, each with a short and a long ambiguous region (\ex{the deer} against \ex{the deer that was brown and graceful}) and, for the optionally transitive verbs, a plausible and an implausible object.
Primes from this set are followed by a fixed set of 48 discovery families as targets, excluding any target that shares a content word with the prime.

A target never shares a content word with its prime, which removes lexical repetition as a source of priming.
In the discovery set each prime is paired with 78 to 92 targets.
The $L$ measure compares two strings that contain the same words in the same order and differ only in the position of the comma.
The $S$ measure scores only the first word of the main clause, so its value does not depend on the length of the continuation.

\section{Statistics}
\label{app:stats}

\begin{figure*}[t]
\centering
\includegraphics[width=\textwidth]{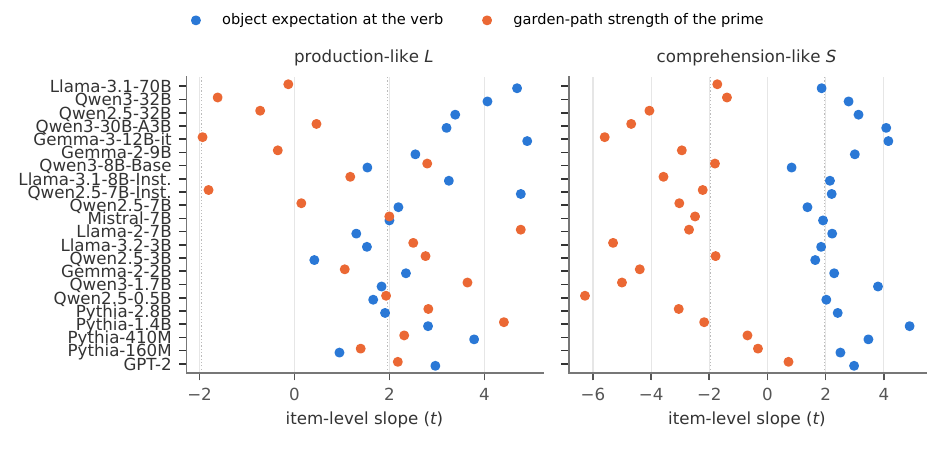}
\caption{Item-level predictors of the residue on the discovery set. Each point is the $t$ value of one predictor in a regression of the family-level contrast $A$ on the model's object expectation after the ambiguous verb (blue) and on the garden-path strength of the prime, the surprisal of its disambiguating word without versus with the comma (orange). Dotted lines mark $|t|=1.96$. Object expectation raises $A$ on both measures. Garden-path strength lowers it on the comprehension-like measure $S$ in 21 of 22 models.}
\label{fig:slopes}
\end{figure*}

For a model and a contrast, we first average the measure over targets for every prime family and condition, then average over prime families.
Intervals resample prime families and target families independently with replacement (1,000 draws) and recompute the contrast, which treats both as random samples of items.
Item-level regressions (Figure~\ref{fig:slopes}) regress the family-level contrast on the two standardized predictors with heteroskedasticity-consistent standard errors.
Across models we pool per-model estimates with the random-effects estimator of \citet{dersimonian1986}, taking the standard error of each model's estimate from its bootstrap interval.
Models are not a random sample of language models, so we read the pooled value as the typical model of our panel and always report how many models show the effect.
Table~\ref{tab:residue} gives the residue of every model on the three item sets, and Figure~\ref{fig:slopes} the item-level regressions of \S\ref{sec:twoforces}.

\begin{table}[t]
\centering\footnotesize
\setlength{\tabcolsep}{2pt}
\begin{adjustbox}{max width=\columnwidth}
\begin{tabular}{lccc}
\toprule
Model & Discovery & Held-out & Christianson\\
\midrule
GPT-2 & 0.16 \ci{0.08}{0.24} & 0.16 \ci{0.08}{0.25} & 0.14 \ci{0.05}{0.22}\\
Pythia-160M & 0.01 \ci{$-$0.18}{0.20} & 0.10 \ci{$-$0.13}{0.33} & 0.27 \ci{0.04}{0.48}\\
Pythia-410M & 0.39 \ci{0.28}{0.52} & 0.25 \ci{0.15}{0.36} & 0.50 \ci{0.38}{0.64}\\
Pythia-1.4B & 0.45 \ci{0.32}{0.59} & 0.32 \ci{0.22}{0.43} & 0.54 \ci{0.41}{0.68}\\
Pythia-2.8B & 0.48 \ci{0.32}{0.63} & 0.45 \ci{0.33}{0.59} & 0.61 \ci{0.47}{0.76}\\
Qwen2.5-0.5B & 0.86 \ci{0.68}{1.04} & 0.53 \ci{0.39}{0.70} & 1.00 \ci{0.81}{1.19}\\
Qwen3-1.7B-Base & 1.16 \ci{0.91}{1.41} & 0.68 \ci{0.49}{0.88} & 1.31 \ci{1.05}{1.58}\\
Gemma-2-2B & 0.65 \ci{0.47}{0.84} & 0.34 \ci{0.19}{0.50} & 1.03 \ci{0.80}{1.26}\\
Qwen2.5-3B & 0.27 \ci{0.15}{0.40} & 0.27 \ci{0.12}{0.43} & 0.30 \ci{0.19}{0.42}\\
Llama-3.2-3B & 0.82 \ci{0.62}{1.03} & 0.26 \ci{0.11}{0.43} & 1.13 \ci{0.87}{1.37}\\
Llama-2-7B & 0.84 \ci{0.65}{1.04} & 0.33 \ci{0.21}{0.45} & 1.07 \ci{0.86}{1.27}\\
Mistral-7B & 0.86 \ci{0.64}{1.10} & 0.39 \ci{0.25}{0.55} & 1.19 \ci{0.96}{1.42}\\
Qwen2.5-7B & 0.90 \ci{0.67}{1.16} & 0.52 \ci{0.33}{0.73} & 1.30 \ci{0.99}{1.59}\\
Qwen2.5-7B-Instruct & 1.11 \ci{0.77}{1.44} & 0.38 \ci{0.07}{0.71} & 1.50 \ci{1.08}{1.92}\\
Llama-3.1-8B-Instruct & 1.09 \ci{0.87}{1.36} & 0.63 \ci{0.43}{0.84} & 1.64 \ci{1.32}{1.96}\\
Qwen3-8B-Base & 0.98 \ci{0.70}{1.26} & 0.28 \ci{$-$0.08}{0.61} & 1.04 \ci{0.76}{1.31}\\
Qwen3-8B & -- & -- & 1.68 \ci{1.35}{2.01}\\
Gemma-2-9B & 0.55 \ci{0.37}{0.76} & 0.18 \ci{0.06}{0.30} & 0.92 \ci{0.71}{1.11}\\
Gemma-3-12B-it & 1.08 \ci{0.75}{1.44} & 0.50 \ci{0.22}{0.82} & 1.84 \ci{1.44}{2.24}\\
Qwen2.5-14B-Instruct & -- & -- & 1.29 \ci{0.93}{1.66}\\
Qwen3-30B-A3B & 1.13 \ci{0.84}{1.45} & 0.49 \ci{0.27}{0.79} & 1.60 \ci{1.22}{1.96}\\
Qwen2.5-32B & 0.61 \ci{0.40}{0.83} & 0.52 \ci{0.31}{0.75} & 1.03 \ci{0.78}{1.25}\\
Qwen3-32B & 0.78 \ci{0.53}{1.07} & 0.33 \ci{0.12}{0.58} & 1.37 \ci{1.03}{1.69}\\
Llama-3.1-70B & 0.52 \ci{0.34}{0.71} & 0.37 \ci{0.23}{0.52} & 0.92 \ci{0.71}{1.12}\\
 
\bottomrule
\end{tabular}
\end{adjustbox}
\caption{Residue $A_L$ (nats) with 95\% two-way bootstrap intervals on the three item sets. Dashes mark models not run on a set.}
\label{tab:residue}
\end{table}

\section{Pre-specified Predictions and Outcomes}
\label{app:prereg}

Each confirmatory experiment was preceded by a dated record of its design, predictions and analysis, committed to version control before any of its data were collected.
Table~\ref{tab:prereg} lists every prediction and its outcome.
Three experiments did not confirm what we expected.
On the held-out families, the between-family relation between object expectation and the residue was measured on $S$ and was confirmed in only 6 of 15 models, whereas the error relation was confirmed in 9 of 15.
Figure~\ref{fig:slopes} shows that object expectation predicts the residue on both measures within the discovery set, and the held-out result indicates that the relation is weaker on $S$ when the families come from sources that differ in verb class.
For NP/S garden paths neither force appeared, which led to the scope statement of \S\ref{sec:twoforces}.
The context manipulation of Appendix~\ref{app:e014} failed its manipulation check.

The within-family manipulation of the error used continuations generated to follow each fixed prefix and checked for grammaticality.
Each family had between four and eight continuations whose disambiguating words differed in surprisal.
The effect of the error remained when only human-written continuations were used, when a naturalness score of the continuation was added as a covariate, and when continuations beginning with an auxiliary were removed ($t$ between $-2.8$ and $-6.2$ in the three models with an error effect).

\begin{table*}[t]
\centering\small
\setlength{\tabcolsep}{4pt}
\begin{tabular}{p{2.6cm}p{6.6cm}p{5.6cm}}
\toprule
Experiment & Pre-specified prediction & Outcome\\
\midrule
Held-out families & (i) object expectation predicts the residue on $S$ (ii) error predicts it with the sign found in discovery (iii) sign of the net residue & (i) 7/17 \emph{(not confirmed)} (ii) 11/17, same sign in 16/17 (sign test $p=.0001$) (iii) 5/9\\
Within-family error & more surprising disambiguation lowers the residue in the models with a discovery error effect & confirmed in 3/3, and null in the 2 models predicted null\\
NP/S & the two forces generalize to NP/S garden paths & \emph{not confirmed} (1/5 and 0/4)\\
MV/RR & error and commitment relations in main-verb/reduced-relative garden paths & (i) error 6/9 (ii) commitment 0/8\\
Case (\cond{nom}) & (i) object expectation predicts the residue without punctuation (ii) error relation negative (iii) comma part of the predictor vanishes & (i) 22/22 (ii) 20/22 (iii) \emph{not confirmed} (reinterpreted as part of object expectation)\\
Measures & (i) residue on $L$ positive (ii) error relation stronger on $S$ than on $L$ & (i) 16/16 (ii) 14/16\\
Passive (\cond{pass}) & passive residue small relative to \cond{gp} & median ratio 0.12 (confirmed)\\
Sampled text & (i) residue positive (ii) raw \cond{gp}$-$\cond{cue} contrast positive (iii) style alignment (iv) size follows $A_L$ & (i) 11/12 (ii) \emph{4/12} (iii) 12/12 (iv) $\rho=0.81$\\
Patching & (i) NP retains $\le 50\%$ and less than MC on $L$ (ii) MC changes $S$ more than NP & (i) 7/9 (ii) 8/9\\
Context & manipulation check on object expectation at the verb & \emph{not met} (0/11)\\
\bottomrule
\end{tabular}
\caption{Pre-specified predictions and outcomes. Counts are models in which the prediction held. Outcomes that did not confirm the prediction are in italics.}
\label{tab:prereg}
\end{table*}

\section{Patching}
\label{app:patching}

For a prime with noun-phrase tokens $N$ and main-clause tokens $M$, we run the prime twice.
The original run keeps all states.
The counterfactual run overwrites, at the output of every layer, the residual-stream vectors at $N$ with the vectors of the same tokens from the donor prime, and recomputes all positions after $N$ with these vectors in place.
Since the models use rotary position embeddings, the residual stream carries no absolute position, and the one-token offset introduced by the donor's comma does not need correction.
The target is then scored against a key--value cache that takes the counterfactual run's keys and values at the positions of the chosen span and the original run's elsewhere.
The prediction of the target's first token comes from the run that supplies the prime's last token.
Table~\ref{tab:patching} gives all estimates with intervals.
In the depth analysis the overwrite starts at layer $\lfloor fL \rfloor$ for $f \in \{0.25, 0.5, 0.75\}$, so that the noun phrase's keys and values below that layer keep the object analysis.
The retained fractions when the overwrite starts at a quarter, half and three quarters of the depth are 0.51 (0.41, 0.61), 1.07 (1.01, 1.12) and 1.01 (1.00, 1.01) for Llama-3.1-8B-Instruct, 0.56 (0.45, 0.65), 1.06 (1.02, 1.12) and 1.02 (1.01, 1.04) for Mistral-7B, and 0.22 (0.02, 0.35), 0.44 (0.28, 0.55) and 1.04 (1.01, 1.08) for Qwen2.5-7B.
Replacing both the noun phrase and the main clause retains 0.32 to 0.67 of the residue, more than replacing the noun phrase alone, because it also removes the adaptation.
For the role direction we take, at each layer, the difference between the mean noun-phrase states of object primes without a comma (\cond{tb}) and subject primes without a comma (\cond{nb}), fit without the prime's own family.
In the middle half of the layers the projection of the noun-phrase states on this direction is set to the subject mean, and the same shift along three norm-matched random directions serves as control.
The residue retained is 0.50 to 0.92 with the role direction and 0.42 to 0.83 with random directions, and the role direction retains less in Llama-3.2-3B, Llama-3.1-8B-Instruct and Qwen2.5-32B.
A direction fit on object primes against subject primes with a comma retains 0.31 to 0.74, less than random directions in all seven models, but it also separates primes with and without a comma.

\begin{table*}[t]
\centering\small
\setlength{\tabcolsep}{4pt}
\begin{tabular}{lccccc}
\toprule
Model & $A_L$ & $\rho$, NP & $\rho$, MC & $\Delta A_S$, NP & $\Delta A_S$, MC\\
\midrule
Pythia-1.4B & 0.46 \ci{0.33}{0.59} & 0.65 \ci{0.56}{0.73} & 0.85 \ci{0.76}{0.92} & $-$0.26 \ci{$-$0.30}{$-$0.22} & 0.22 \ci{0.11}{0.34}\\
Pythia-2.8B & 0.49 \ci{0.34}{0.65} & 0.50 \ci{0.41}{0.59} & 0.74 \ci{0.64}{0.83} & $-$0.35 \ci{$-$0.42}{$-$0.29} & 0.40 \ci{0.25}{0.55}\\
Llama-3.2-3B & 0.82 \ci{0.62}{1.03} & 0.39 \ci{0.30}{0.48} & 0.83 \ci{0.72}{0.95} & $-$0.51 \ci{$-$0.60}{$-$0.42} & 1.12 \ci{0.92}{1.33}\\
Mistral-7B & 0.86 \ci{0.64}{1.09} & 0.40 \ci{0.28}{0.49} & 0.89 \ci{0.74}{1.10} & $-$0.55 \ci{$-$0.66}{$-$0.46} & 0.93 \ci{0.74}{1.11}\\
Qwen2.5-7B & 0.90 \ci{0.66}{1.16} & 0.14 \ci{$-$0.05}{0.27} & 1.34 \ci{1.12}{1.71} & $-$0.84 \ci{$-$0.99}{$-$0.71} & 1.13 \ci{0.93}{1.33}\\
Llama-3.1-8B-Inst. & 1.09 \ci{0.87}{1.36} & 0.28 \ci{0.17}{0.38} & 1.03 \ci{0.91}{1.17} & $-$0.83 \ci{$-$0.96}{$-$0.71} & 1.33 \ci{1.11}{1.55}\\
Qwen3-8B-Base & 0.95 \ci{0.68}{1.22} & 0.71 \ci{0.55}{0.92} & 0.96 \ci{0.80}{1.16} & $-$1.11 \ci{$-$1.32}{$-$0.95} & 2.00 \ci{1.67}{2.34}\\
Qwen2.5-32B & 0.61 \ci{0.40}{0.83} & 0.01 \ci{$-$0.35}{0.21} & 1.94 \ci{1.55}{2.63} & $-$0.72 \ci{$-$0.83}{$-$0.60} & 1.55 \ci{1.32}{1.79}\\
Llama-3.1-70B & 0.52 \ci{0.34}{0.71} & 0.01 \ci{$-$0.36}{0.22} & 1.89 \ci{1.47}{2.63} & $-$0.79 \ci{$-$0.93}{$-$0.66} & 1.46 \ci{1.21}{1.69}\\
 
\bottomrule
\end{tabular}
\caption{Patching results with 95\% two-way bootstrap intervals. $\rho$: retained fraction of the residue on $L$. $\Delta A_S$: change of the residue on $S$ in nats. NP and MC: the target reads counterfactual states at the noun phrase or at the main clause.}
\label{tab:patching}
\end{table*}

\section{Context Manipulation}
\label{app:e014}

Each of 87 discovery families received five context sentences, written by Qwen3-32B from a fixed instruction and checked automatically for length, the absence of commas, and the absence of the family's verbs and main-clause words.
Two contexts present the noun phrase as the subject's goal (\ex{The man had been tracking the deer since dawn.}), two present the subject and the noun phrase as doing unrelated things (\ex{The man studied the map while the deer drifted downstream.}), and one only places them together.
Six families for which no context passed the checks were dropped as specified.
Each context precedes all four primes of Equation~\ref{eq:A}, so the context's direct effect on the target cancels.

The specified manipulation check required object expectation after the ambiguous verb to be higher after goal contexts in more than 70\% of families.
It rose on average in every model but in only 48\% to 64\% of families, so the confirmatory test is uninformative by its own criterion.
The contexts did change the garden path itself: the surprisal of the disambiguating word without versus with the comma rose by 0.9 to 1.7 nats (Table~\ref{tab:e014}).
This analysis was added after the first two models had been scored.
Under it, the residue on $S$ is larger after goal contexts in 10 of 11 models, which runs against an account in which the prime acts only through its disambiguation error, since that error also grows.

\begin{table}[t]
\centering\footnotesize
\setlength{\tabcolsep}{2.5pt}
\begin{adjustbox}{max width=\columnwidth}
\begin{tabular}{lccccc}
\toprule
Model & $\Delta$exp. & fam. & $\Delta$GP & $\Delta A_L$ & $\Delta A_S$\\
\midrule
Pythia-1.4B & 0.51 & 64\% & 0.88 & $-$0.21 (0.03) & $-$0.03 (0.03)\\
Pythia-2.8B & 0.39 & 63\% & 1.14 & $-$0.10 (0.02) & 0.06 (0.03)\\
Gemma-2-2B & 0.16 & 53\% & 1.29 & $-$0.06 (0.04) & 0.06 (0.05)\\
Llama-3.2-3B & 0.17 & 48\% & 1.06 & $-$0.06 (0.04) & 0.12 (0.04)\\
Mistral-7B & 0.30 & 57\% & 1.35 & 0.13 (0.06) & 0.28 (0.05)\\
Qwen2.5-7B & 0.26 & 52\% & 1.10 & 0.29 (0.07) & 0.27 (0.06)\\
Qwen2.5-7B-Instruct & 0.29 & 51\% & 1.41 & 0.46 (0.11) & 0.51 (0.10)\\
Llama-3.1-8B-Instruct & 0.28 & 59\% & 1.05 & $-$0.02 (0.05) & 0.36 (0.06)\\
Qwen3-8B-Base & 0.58 & 59\% & 1.73 & 0.13 (0.10) & 0.08 (0.08)\\
Gemma-2-9B & 0.29 & 53\% & 1.72 & $-$0.01 (0.05) & 0.16 (0.07)\\
Llama-3.1-70B & 0.37 & 57\% & 1.40 & 0.26 (0.05) & 0.32 (0.06)\\
 
\bottomrule
\end{tabular}
\end{adjustbox}
\caption{Context manipulation, goal minus unrelated contexts. $\Delta$exp.: object expectation at the verb (pre-specified check), and the share of families in which it rose. $\Delta$GP: garden-path strength. $\Delta A_L$, $\Delta A_S$: change of the residue, with standard errors over families.}
\label{tab:e014}
\end{table}

\section{Overridden Statements}
\label{app:facts}

The 200 items combine ten frames (meeting day, location of objects, city of residence, start time, favourite colour, due month, profession, material, instrument, language) with five entities and two pairs of values in both orders.
Every context starts with a short label (\ex{Note:}, \ex{Mistake:}), so that the overridden statement is tokenized identically in all conditions.
The measure is $M=\log P(A)-\log P(B)$ after the probe, and $R=M_{\text{CORR}}-M_{\text{CTRL}}$.
In the re-encoding test, the residual-stream states of the overridden statement's tokens in CORR are replaced at every layer by those of the same tokens in PRE, where \ex{Mistake:} precedes the statement, and the same replacement in CTRL is subtracted.
This leaves the pull largely intact (Table~\ref{tab:facts}), as the pre-specified prediction did not expect.
The knockout analysis uses full sequences with a four-dimensional attention mask that blocks the listed query--key pairs in every layer and head.
Retained fractions of the pull, for blocking the answer's access to the statement, the answer's access to the value token only, and all later positions' access to the statement, are 1.10, 0.94, 0.12 (Llama-3.2-3B); 1.64, 1.53, 0.21 (Llama-3.1-8B-Instruct); 1.34, 1.10, $-$0.06 (Mistral-7B); 2.08, 1.70, $-$0.03 (Qwen3-8B-Base); 1.12, 1.11, 0.16 (Gemma-2-9B); 0.16, $-$0.04, 1.13 (Qwen2.5-7B); 0.48, 0.66, $-$0.20 (Llama-3.1-70B); and $-$3.02, $-$3.39, 2.90 (Qwen2.5-32B, whose small pull of 0.56 nats reverses).

\begin{table}[t]
\centering\footnotesize
\setlength{\tabcolsep}{3pt}
\begin{adjustbox}{max width=\columnwidth}
\begin{tabular}{lcccc}
\toprule
Model & $R$ & CORR$-$PRE & $\rho$ CLAIM & $\rho$ FULL\\
\midrule
Llama-3.2-3B & 1.65 \ci{1.53}{1.77} & 0.53 & 1.00 & 0.72\\
Llama-3.1-8B-Instruct & 1.79 \ci{1.65}{1.95} & 2.04 & 0.72 & $-$0.16\\
Mistral-7B & 2.17 \ci{1.97}{2.37} & 0.14 & 0.81 & 0.55\\
Qwen2.5-7B & 1.35 \ci{1.23}{1.47} & 0.41 & 0.98 & 0.57\\
Qwen2.5-7B-Instruct & $-$2.23 \ci{$-$2.57}{$-$1.90} & $-$1.17 & 1.13 & 1.86\\
Qwen3-8B-Base & 2.26 \ci{2.06}{2.46} & 0.55 & 1.00 & 1.13\\
Gemma-2-9B & 1.57 \ci{1.46}{1.68} & $-$0.01 & 0.86 & 0.33\\
Qwen2.5-32B & 0.56 \ci{0.44}{0.68} & $-$0.42 & 1.08 & 0.13\\
Llama-3.1-70B & 1.88 \ci{1.77}{1.99} & 0.43 & 0.80 & 0.49\\
 
\bottomrule
\end{tabular}
\end{adjustbox}
\caption{Overridden statements (nats). $R$: pull of the overridden value beyond a matched mention. $\rho$: retained fraction of $R$ after re-encoding the overridden statement (CLAIM) or the whole context (FULL).}
\label{tab:facts}
\end{table}

\section{Coding of Sampled Completions}
\label{app:coding}

Sampled continuations of \ex{While}~$s$~$v$ are coded from part-of-speech tags, because a dependency parser is itself misled by comma-less input of exactly this kind.
A continuation that begins with punctuation is intransitive.
One that begins with an object pronoun, or with a noun phrase that is followed by punctuation, a second noun phrase or the end of the clause, is transitive.
A noun phrase followed by a finite verb is intransitive, and if no comma precedes it, the continuation is itself a garden-path sentence.
A second version of the rules also resolves coordination after the noun phrase.
Continuations with a relative clause or an unresolved coordination after the noun phrase are coded as ambiguous and excluded (1.5\% to 13\% per model).
On 1,066 sentences from our materials with known structure the rules are correct on 1,063.
On 180 generated continuations sampled across classes, one annotator judged 27 to 29 of 30 transitive codes and 29 of 30 intransitive codes correct.
The results of \S\ref{sec:residue} and \S\ref{sec:transfer} use the first version, and the second gives the same conclusions with the residue positive in 12 of 12 models.

\end{document}